\documentclass[10pt,twocolumn,letterpaper]{article}
\usepackage{cvpr}
\usepackage{times}
\usepackage{epsfig}
\usepackage{graphicx}
\usepackage{amsmath}
\usepackage{amssymb}
\cvprfinalcopy
\newcommand{\latin}[1]{{\it #1}}
\usepackage{adjustbox}
\usepackage[inline, shortlabels]{enumitem}
\usepackage{fancyvrb}
\usepackage{arydshln}
\usepackage{mathtools, cuted}
\usepackage{breqn}
\usepackage{amsmath}
\usepackage{amssymb}
\usepackage{dsfont}
\usepackage[linesnumbered,ruled]{algorithm2e}
 
\usepackage{graphicx} 
\usepackage{subcaption}
\usepackage{pdfpages}
\pdfoutput=1
\usepackage[pagebackref=true,breaklinks=true,letterpaper=true,colorlinks,bookmarks=false]{hyperref}
\usepackage{float}
\usepackage{anyfontsize}
\newcounter{nbdrafts}
\makeatletter
\newcommand{\checknbdrafts}{
\ifnum \thenbdrafts > 0
\@latex@warning@no@line{**********************************************************************}
\@latex@warning@no@line{* The document contains \thenbdrafts \space draft note(s)}
\@latex@warning@no@line{**********************************************************************}
\fi}

\makeatother

\begin{document}

\title{SGAN: An Alternative Training of Generative Adversarial Networks}
\author{Tatjana Chavdarova \ and \ Fran\c cois Fleuret \\
Idiap Research Institute\\
\'Ecole Polytechnique F\'ed\'erale de Lausanne \\
{\tt\small tatjana.chavdarova@idiap.ch \ francois.fleuret@idiap.ch}
}

\maketitle
\begin{abstract}

The Generative Adversarial Networks (GANs) have demonstrated impressive performance for data synthesis, and are now used in a wide range of computer vision tasks. In spite of this success, they gained a reputation for being difficult to train, what results in a time-consuming and human-involved development process to use them.

We consider an alternative training process, named SGAN, in which several adversarial ``local'' pairs of networks are trained independently so that a ``global'' supervising pair of networks can be trained against them. The goal is to train the global pair with the corresponding ensemble opponent for improved performances in terms of mode coverage. This approach aims at increasing the chances that learning will not stop for the global pair, preventing both to be trapped in an unsatisfactory local minimum, or to face oscillations often observed in practice. To guarantee the latter, the global pair never affects the local ones.

The rules of SGAN training are thus as follows: the global generator and discriminator are trained using the local discriminators and generators, respectively, whereas the local networks are trained with their fixed local opponent.

Experimental results on both toy and real-world problems demonstrate that this approach outperforms standard training in terms of better mitigating mode collapse, stability while converging and that it surprisingly, increases the convergence speed as well.

\end{abstract}

\section{Introduction}\label{sec-intro}
An important research effort has recently focused on improving the convergence analysis of the Generative Adversarial Networks~\cite{GoodfellowGAN2014}.
This family of unsupervised learning algorithms provides powerful generative models, and have found numerous and diverse applications~\cite{IsolaZZE16,LedigTHCATTWS16,NguyenYBDC16,ZhangXLZHWM16}.

\begin{figure}[!ht]
  \centering
  \includegraphics[width=\linewidth]{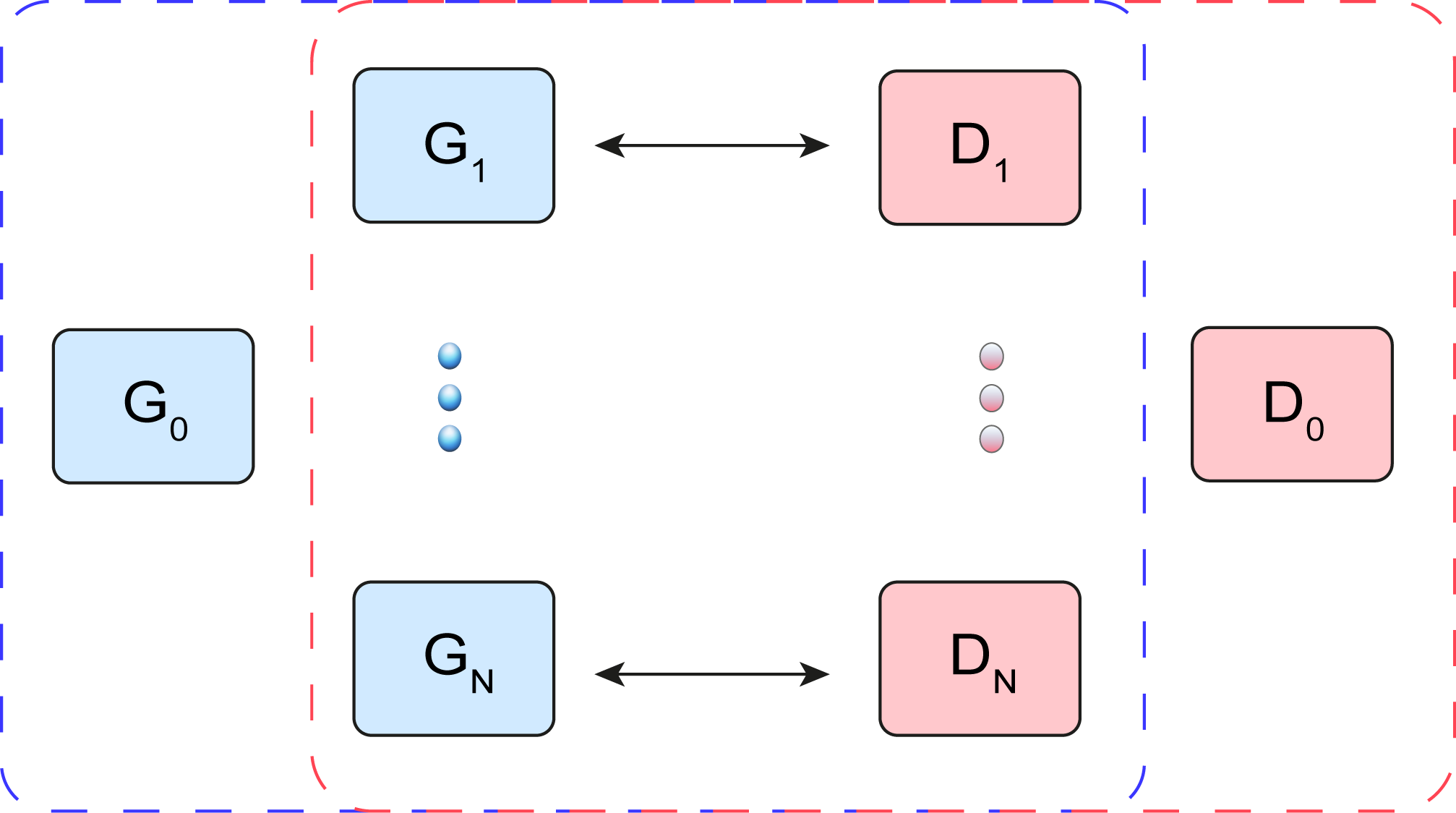}
  \caption{Conceptual illustration of SGAN.
  There are $N{+}1$ pairs, of which the pair ($G_{0}$, $D_{0}$) is not trained directly.
  $D_0$ is trained with $G_i$, $i{=}1, \dots ,N $, and $G_0$ is trained with $D_i$, $i{=}1, \dots ,N$, as illustrated with the dashed line rectangles.}
\label{fig-concept}
\end{figure}

Different from traditional generative models, a GAN generator represents a mapping  $G\!:  z \mapsto x$, such that if $z$ follows a known distribution $p_z$, then $x$ follows the distribution $p_{d}$ of the data.
Notably, this approach omits an explicit representation of $p_g(x)$, or the ability to apply directly a maximum-likelihood maximization for training.
This is aligned with the practical need, which is that we do not need an explicit formulation of $p_g(x)$, but rather a mean to sample from it, preferably in a computationally efficient manner.
The training of the generator includes a discriminative model $D\!: x \mapsto y \in [0,1]$ whose output represents an estimated probability that $x$ originates from the dataset, given that there was probability $0.5$ that was the case and $0.5$ it was generated by $G$.

The two training steps--also referred as bi-level optimization of the GAN algorithm--consist of training $D$ to distinguish real from fake samples, and training $G$ to fool $D$ by generating synthetic samples indistinguishable from real ones.

Hence, these two competing models play the following two-player minimax--alternatively zero-sum--game:

{\setlength{\mathindent}{0cm}
\begin{equation}
\min_{G} \max_{D} \mathop{\mathbb{E}}_{x \sim p_{d}} [\log{D(x)}]{+}\mathop{\mathbb{E}}_{z\sim p_z}[\log(1{-}D(G(z)))].
\end{equation}} \label{eq-minimax}

The two models are parametrized differentiable functions $G(z; \theta_g)$ and $D(x; \theta_d)$, implemented with neural networks, whose parameters $\theta_g$ and $\theta_d$ are optimized iteratively.
In functional space, the competing models are guaranteed to reach a Nash Equilibrium, in particular under the assumptions that we optimize directly $p_g$ instead of $\theta_g$ and that the two networks have enough capacity.
At this equilibria point, $D$ outputs probability $0.5$ for any input.

In practice, GANs are difficult to optimize, and
practitioners have amassed numerous techniques to improve stability of the training process~\cite{dcgan}.
However, the neglected inherited problems of the neural networks such as the lack of convexity, the numerical instabilities of some of the involved operations, the limited representation capacity, as well as the problem of vanishing and exploding gradients often emerge in practice.

As a consequence, current state-of-the-art GAN variants~\cite{arjovsky2017wgan, wgangp} eliminate gradient instabilities, which mitigated the discrepancy between the theoretical requirement that $D$ should be trained up to convergence before updating $G$, and the practical procedures of vanilla GAN for which this is not the case. These results are important, as vanishing or exploding gradients results in $G$ to produce samples of noise.

However, oscillations between noisy patterns and samples starting to look like real data while the algorithm is converging, as well as failures of capturing $p_{d}$, are not resolved.
In addition, in practical applications, it is very difficult to assess the diversity of the generated samples. ``Fake'' samples may look realistic but could be similar to each other -- indicating that the modes of $p_d$ have been only partially ``covered'' by $p_g$. This is a problem that arises and is referred as \textit{mode collapse}.
As a result, a golden rule remains that one does multiple trials of combinations of hyperparameters, architectural and optimization choices, and variants of GANs.
As under different choices the performances vary, this is followed by tedious and subjective assessments of the quality of the generated samples in order to select a generator.

As the root cause, we do not see the algorithm itself, but rather the limited representational capability of the deep neural networks which is most critical in the early optimization phases, and causes an increased number of local saddle points (Nash equilibria) -- what could explain the oscillations observed in practice.
These problems of oscillations, limited representation and the unsatisfying varying performances of the GAN algorithm have not been directly addressed.

As a summary, what made GAN distinctly powerful is the opponent-wise engagement of two networks belonging to an already outperforming class of algorithms.
The more we enforce constraints, either architectural or functional, the better the stability would be. However, the price we pay is a reduced quality of generated samples or convergence speed at the minimum.
In this paper, we raise the question if we could sustain the game-play and yet improve stability, guarantees, or at least consistency in the performances of the algorithm, whose final goal is to produce a good generator.

We propose a novel way of training a \textit{global} pair $(G_0, D_0)$, such that the optimization process will make use of the ``flow of information'' generated by training an ensemble of $N$ adversarial pairs $(G_1, D_1), \dots, (G_N, D_N)$, as sketched in Figure~\ref{fig-concept}.
As we shall discuss, this two-level hierarchy that we will enforce could allow us to later extend it in a way that $G_0$ and $D_0$ can employ strategy against each other.

The rules of this game are as follows:
$G_0$ and $D_0$ can solely be trained with  $\{D_1, \dots D_N\}$ and  $\{G_1, \dots G_N\}$, respectively, and local pairs do not have access to outputs or gradients from $G_0$ and $D_0$.

The most prominent advantages of such a training are:
\begin{enumerate}
\item if the training of a particular pair degrades or oscillates, the global networks continue to learn with higher probability;
\item it is much more likely that training one pair will fail than training all of them, hence the choice of not letting global models to affect the ensemble;
\item if the models' limited capacity is taken into account \latin{i.e.} $p_g$ can capture a limited number of modes of $p_d$ (which increases with the number of training iterations), and under the assumption that each mode of $p_d$ has a non-zero probability of being captured,
then the modeled distribution by the ensemble is closer to $p_d$ in some metric space due to the statistical averaging; and conveniently
\item large chunks of the computation can be carried out in parallel making the time overhead negligible.
\end{enumerate}

In what follows, we first review in \S~\ref{sec-related_work} GAN variants we use in the experimental evaluation of our SGAN algorithm, which to the best of our knowledge are the current state-of-the-art methods.
We then describe SGAN in detail in \S~\ref{sec-method}, and present thorough experimental evaluation in \S~\ref{sec-experiments} as well as in the Appendix.

We then present some methods that although unrelated to the SGAN approach, do propose multi-agent structure in \S~\ref{sec:related-multi-network}.
We finally discuss two viewpoints of SGAN and possible extensions in \S~\ref{ss-SGAN_in_game_theory}.

\section{Related work: Variants of the GAN algorithm}\label{sec-related_work}
Optimizing Eq.~\ref{eq-minimax} amounts to minimizing the Jensen-Shannon divergence between the data and the model distribution $JS(p_g, p_d)$~\cite{GoodfellowGAN2014}. 
More generally, GANs learn $p_d$ by minimizing a particular f-divergence between the real samples and the generated samples~\cite{fgan}.

With a focus on generating images, ~\cite{dcgan} proposes specific architectures of the two models, named Deep Convolutional Generative Adversarial Networks--\textbf{DCGAN}. \cite{dcgan}  also enumerates a series of practical guidelines, critical for the training to succeed. Up to this point, when the architecture and the hyper-parameters are empirically selected, DCGAN demonstrates outperforming results both in terms of quality of generated samples and convergence speed.

To ensure usable gradient for optimization, the mapping $\theta_d \mapsto p_d$ should be differentiable, and to have a non-zero gradient everywhere. As the $JS$ divergence does not take into account the Euclidean structure of the space, it may fail to make the optimization move distributions closer to each other if they are ``too far apart''~\cite{arjovsky2017wgan}. Hence, ~\cite{arjovsky2017wgan} suggests the use of the Wasserstein distance, which precisely accounts for the Euclidean structure. Through the Kantorovich Rubinstein duality principle~\cite{villani2009}, this boils down to having a $K$-Lipschitz discriminator.

From a purely practical standpoint, this means that strongly regularizing the discriminator prevents the gradient from vanishing through it, and helps the optimization of the generator by providing it with a long-range influence that translates into a non-zero gradient.

In \textbf{WGAN} ~\cite{arjovsky2017wgan} the Lipschitz continuity is forced through weight clipping, which may make the optimization of $D$ harder--as it makes the gradient with respect to $D$'s parameters vanish--and often leads to degrading the overall convergence.
It was later proposed to enforce the Lipschitz constraint smoothly by adding a term in the loss which penalizes gradients whose norm is higher than one--\textbf{WGAN-GP}~\cite{wgangp}.

Motivated by game theory principles,~\cite{dragan} derives combined solution of vanilla GAN and WGAN with gradient penalty.
In particular, the authors aim at smoothing the value function via regularization by minimizing the regret over the training period, so as to mitigate the existence of the multiple saddle points. 
Finally, while building on vanilla GAN, the proposed algorithm named \textbf{DRAGAN}--Deep Regret Analytic GAN--forces the constraint on the gradients of $D(x)$ solely in local regions around real samples.

\begin{algorithm}[ht!]
\small
  \SetKwInOut{Input}{Input}
  \SetKwInOut{Output}{Output}

  \Input{ $\mathcal{X}_{inf}$, $N$, I, $I_D$ }

  $\mathcal{G}$, $\mathcal{D}$ = {\tt init}($N$)\; \label{alg:strategic:init}
  $G_0$, $D_0$ = {\tt init}($1$) \label{alg:strategic:init2}

  \For{$i \in \{1 \dots I\}$}{
    \For{$n \in \{1 \dots N\}$}{ \label{alg:strategic:local}

		\For{$j \in \{1 \dots I_D\}$}{ \label{alg:strategic:localDtrain}
				{\tt zeroGradients}($\mathcal{D}[n]$)\; \label{alg:strategic:zeroD}
				{\tt backprop}($\mathcal{G}[n], \mathcal{D}[n], \mathcal{X}_{inf}$)\;
				{\tt updateParameters}($\mathcal{D}[n]$)\; \label{alg:strategic:updateD}
		}
        {\tt zeroGradients}($\mathcal{G}[n]$)\; \label{alg:strategic:zeroG}
        {\tt backprop}($\mathcal{G}[n], \mathcal{D}[n], \mathcal{X}_{inf}$)\;
		{\tt updateParameters}($\mathcal{G}[n]$)\; \label{alg:strategic:updateG}
    }
    $\mathcal{D}^{msg}$ = {\tt copy}($\mathcal{D}$)\; \label{alg:strategic:Dcpy}

    \For{$n \in \{1 \dots N\}$}{\label{alg:strategic:trDmsg}
		\For{$j \in \{1 \dots I_D\}$}{ \label{alg:strategic:DMSG}
			{\tt zeroGradients}($\mathcal{D}^{msg}[n]$)\; \label{alg:strategic:zeroDMSG}
      		{\tt backprop}($G_0, \mathcal{D}^{msg}[n], \mathcal{X}_{inf}$)\;
      		{\tt updateParameters}($\mathcal{D}^{msg}[n]$)\; \label{alg:strategic:updateDMSG}
    	}
    }\label{alg:strategic:Dcpy-done}

    {\tt zeroGradients}($G_0$)\; \label{alg:strategic:zeroG0}
    \For{$n \in \{1 \dots N\}$}{
      {\tt backprop}($G_0, \mathcal{D}^{msg}[n], \mathcal{X}_{inf}$)\;
    }
    {\tt updateParameters}($G_0$)\; \label{alg:strategic:updateG0}

    {\tt zeroGradients}($D_0$)\; \label{alg:strategic:zeroD}
    \For{$n \in \{1 \dots N\}$}{
      {\tt backprop}($\mathcal{G}[n], D_0, \mathcal{X}_{inf}$)\;
    }
    {\tt updateParameters}($D_0$)\; \label{alg:strategic:updateD0}

  }
  \Output{$G_0, D_0$}
  \caption{Pseudocode for SGAN.}
  \label{alg:strategic}
\end{algorithm}

\section{Method}\label{sec-method}

\paragraph{Structure.}
We use a set $\mathcal{G}=\{G_1, \dots G_N\}$ of $N$ generators, a set $\mathcal{D}=\{D_1 \dots D_N\}$ of $N$ discriminators, and a global
generator-discriminator pair $(G_0, D_0)$, as sketched in Figure~\ref{fig-concept}.

\paragraph{Summary of a simplified-SGAN implementation.}
The pairs $(G_n, D_n), \ n = 1, \dots N$ are trained individually in a standard approach.
In parallel to their training, $D_0$ is optimized to detect samples generated by any of the local generators $G_1, \dots, G_N$, and similarly $G_0$ is optimized to fool any of the local discriminators $D_1, \dots, D_N$.

Note that, to satisfy the theoretical analyses of minimizing the Wasserstein distance and the Jensen-Shannon divergence for WGAN and GAN, respectively, the above procedure of training implies that each $\{D_1 \dots D_N\}$ \textit{should be trained with $G_0$} at each iteration of SGAN. Solely by following such a procedure $G_0$  follows the principles of the GAN framework~\cite{GoodfellowGAN2014}, which trains it with gradients ``meaningful'' for it.

\paragraph{Introducing ``messengers'' discriminators for improved guarantees.}
To prevent that one of the network pairs ``influences'' the ensemble, and thus keep the guarantees of successful training, we propose to train $G_0$ against herein referred as ``messengers'' discriminators $D^{msg}_1, \dots, D^{msg}_N$, which at re-created at every iteration as clones of $D_1, \dots, D_N$, optimized against $G_0$.

We empirically observed that this strategy helps consecutive steps to be more coherent, and improves drastically the convergence. It is worth noting that, despite the increased complexity in terms of obtaining the theoretical analyses, such an approach is practically convenient since it allows for training $G_0$ in parallel to the local pairs.

\subsection{Description of SGAN}

More formally, let $\mathcal{X}_{inf}$ be a sampling operator over the dataset, which provides mini-batches of i.i.d. samples $x{\sim}p_d$.

Let \textit{backprop} be a function that given a pair $G$ and $D$, buffers the  updates of the networks' parameters, \textit{updateParameters} that actually updates the parameters using these buffers, and \textit{zeroGradients} resets these buffers. Also, let \textit{init} be a function that initializes a given number of pairs of $G$ and $D$. Let $N$ be the number of pairs to be used.
The algorithm iterates for a given number of iterations $I$, and depending on the used GAN variant, each discriminator network is updated either once or several times, hence the $I_D$ parameter.

At each iteration, foremost the local models are being updated (line \ref{alg:strategic:local}).

Then, to obtain meaningful gradients for $G_0$, without affecting the
local models, we first make a copy of the latter (line
\ref{alg:strategic:Dcpy}) into the ``messenger discriminators'' $\mathcal{D}^{msg}$, and update them against $G_0$ (lines~\ref{alg:strategic:trDmsg} -~\ref{alg:strategic:Dcpy-done}).  We then update $G_0$ jointly versus all
of the discriminators (lines \ref{alg:strategic:zeroG} -~\ref{alg:strategic:updateG}).

As $D_0$ does not affect generators it is trained with, it is directly
updated jointly versus all of the local generators (lines~\ref{alg:strategic:zeroD} -~\ref{alg:strategic:updateD0}).

Note that for clarity in Alg.~\ref{alg:strategic} we
present SGAN sequentially. However, each iteration of the training can be parallelized since $G_0$ is trained with a copy of $\mathcal{D}$, and the local pairs can be trained independently.  In addition, Alg.~\ref{alg:strategic} can
be used with different GAN variants.

SGAN can also be implemented with weight-sharing (see \S~\ref{sec:exp}) since low-level features can be learned jointly across the networks. As of this reason as well as for clearer insights on extensions of SGAN, we do not omit $D_0$ from Alg.~\ref{alg:strategic}. In addition, whether the discriminator can be made use of is not a closed topic. In fact, a recent work answers  in the affirmative~\cite{NIPS2017_6639}.

\section{Experiments}\label{sec-experiments}
\paragraph{Datasets.}\label{par-datasets}

As toy problems in $\mathds{R}^2$ we used
\begin{enumerate*}[series = tobecont, itemjoin = \quad, label=(\roman*)]
\item mixtures of $M$ Gaussians ($M$-GMM) whose means are regularly positioned either on a circle or a grid, with $M=8, 10, \, \text{or} \ 25$, and 
\item the classical Swiss Roll toy dataset~\cite{Marsland:2009:MLA:1571643}.
\end{enumerate*}
In the former case, we manually generate such datasets, by using a mixture of $M$ Gaussians with modes that are uniformly distributed in a circle or in a grid. 
With such an evaluation, we follow related works--for \latin{e.g} \cite{wgangp, dragan, 2017arXivAdaGAN} since GANs in prior work often failed to converge even on such simplistic datasets \cite{unrolledgan}.

To assess SGAN or real world applications, we used:
\begin{enumerate}
\item small scale datasets: CIFAR10~\cite[chapter 3]{cifar10},
MNIST~\cite{mnistlecun}, as well as the recent FASHION-MNIST~\cite{fashionmnist};
\item  large scale datasets: CelebA~\cite{celebA}, LSUN~\cite{yu15lsun} using its ``bedroom'' class, and ImageNet~\cite{ILSVRC15}; as well as
\item large language corpus of text in English, known as One Billion Word Benchmark~\cite{jozefowicz2016exploring}.
\end{enumerate}

\paragraph{Methods.}\label{par-methods}
As WGAN with gradient penalty~\cite{wgangp} outperformed WGAN with weight clipping~\cite{arjovsky2017wgan} in our experiments, herein as  ``WGAN'' we refer to the former. Similarly, we may use GAN and DRA as an abbreviation of DCGAN and DRAGAN, respectively.
For conciseness, let us adopt the following notation regarding SGAN: we prefix the type of GAN with $N$-S, where $N$ is the number of local pairs being used. For example, SGAN with 5  WGAN local pairs and one global WGAN pair would be denoted as 5-S-WGAN.

\paragraph{Implementation.}
For experiments on toy datasets, we used separate $2{\times}(N{+}1)$ neural networks. Regarding experiments on real-world image datasets, it is reasonable to learn the low-level features such as edges jointly across the networks. We run both types of experiments i.e. using separate networks, as well as with sharing parameters. In the latter case, in our implementation, we used approximately half of the parameters to be shared among the generators, and analogously same quantity among the discriminators. For further details on our implementation, see Appendix. 

As a deep learning framework we used PyTorch~\cite{pytorch}.

\begin{figure*}[!htb]
    \centering
    \begin{subfigure}[t]{0.23\linewidth}
        \centering
        \includegraphics[width=\linewidth,trim={1.5cm 1cm 1cm 1.5cm}, clip]{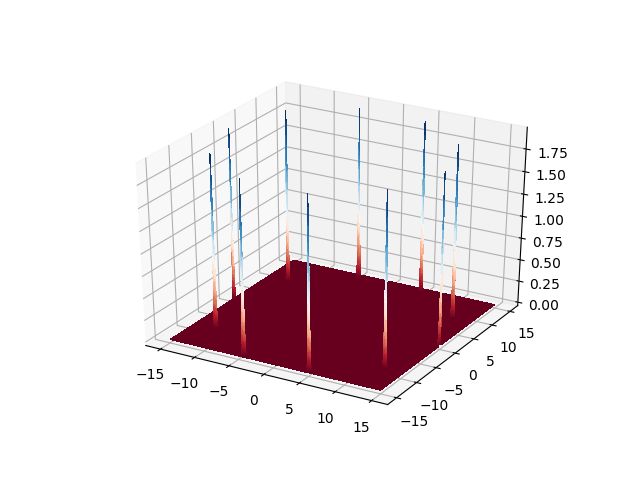}
        \caption{Real data (10-GMM)}\label{subfig-real_data}
    \end{subfigure} 
    \begin{subfigure}[t]{0.23\linewidth}
        \centering
        \includegraphics[width=\linewidth,trim={1.5cm 1cm 1cm 1.5cm}, clip]{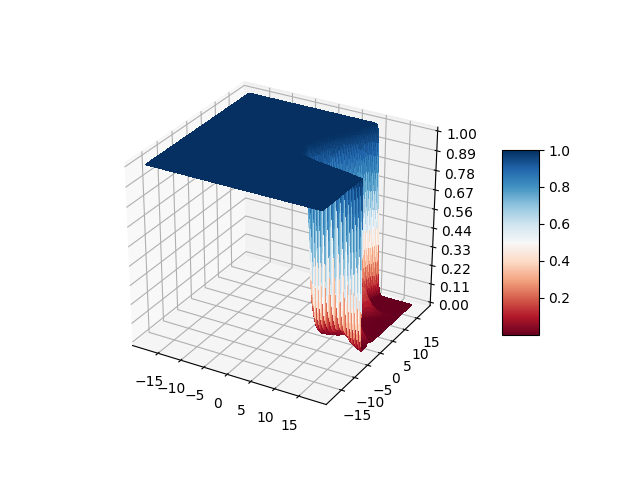}
        \caption{Discriminator output}\label{subfig-direct_disc}
    \end{subfigure}
    \begin{subfigure}[t]{0.23\linewidth}
        \centering
        \includegraphics[width=\linewidth,trim={1.5cm 1cm 1cm 1.5cm}, clip]{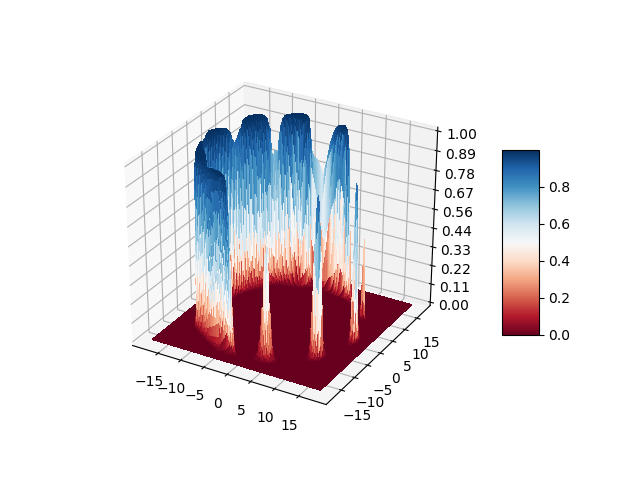} 
        \caption{S-Discriminator output}\label{subfig-indirect_disc}
    \end{subfigure}
    \begin{subfigure}[t]{0.26\linewidth}
        \centering
        \includegraphics[width=\linewidth]{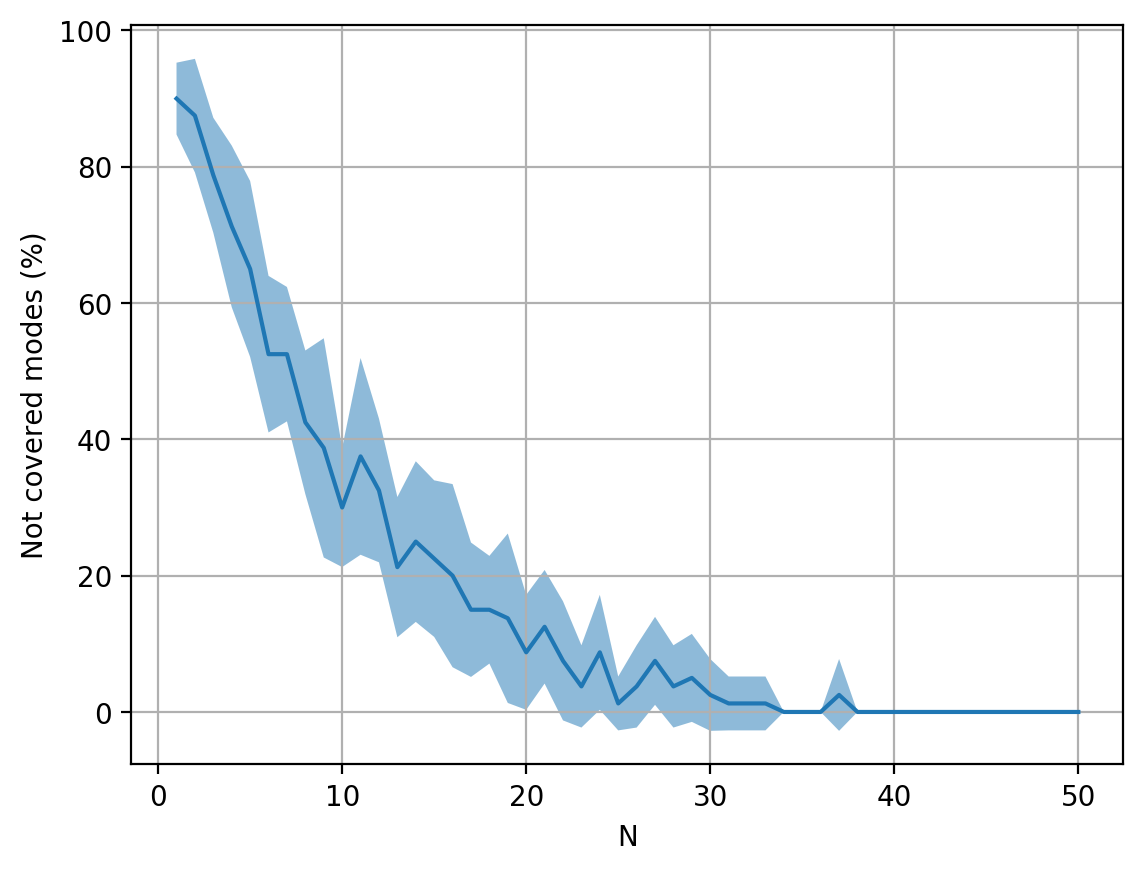} 
        \caption{Not-covered modes (\%)}\label{subfig-modes_not_cov}
    \end{subfigure}
    \caption{Figures (\subref{subfig-real_data}-\subref{subfig-indirect_disc}) depict a toy experiment with vanilla GAN. Figure (\subref{subfig-modes_not_cov}) depicts the percentage of not covered modes (y-axis) by the generators, as more pairs are used (x-axis). See text for details, \S~\ref{sec:exp}. }
	\label{fig-ensemblegan}
\end{figure*}

\paragraph{Metrics.}
A serious limitation to improve GANs is the lack of a proper means of evaluating them. When dealing with images, the most commonly used measure is the so-called \textbf{Inception score}~\cite{Salimans2016improvingGANs}. This metric feeds a pre-trained Inception model~\cite{inceptionmodel} with generated images and measures the KL divergence between the predicted conditional label distribution and the actual class probability distribution.
The mode collapse failure is reflected by the mode's class being less frequent, making the conditional label distribution more deterministic.

Although it was shown to correlate well with human evaluation on CIFAR10~\cite{Salimans2016improvingGANs}, in practice, there are cases where it does not provide a consistent performance estimate~\cite{is2}. In particular, we tested real data images of ImageNet, LSUN-bedroom, CIFAR10, and CelebA, and we obtain the following scores:
46.99 (3.547), 2.37 (0.082), 10.38 (0.502), 2.50 (0.082), respectively for each of the datasets. The high variance of it on real data samples, suggests that utilizing classifier specifically trained for that dataset may improve this performance estimate. 
Thus, we adopt it as is for CIFAR10 as it turned into a standard way of evaluating GANs, whereas for MNIST we utilize a classifier specifically trained on it. 
In the former case, we use the original implementation of it~\cite{Salimans2016improvingGANs} and a sample of $p_g$  of size $50{\cdot}10^3$, whereas for the latter we use our own implementation in PyTorch~\cite{pytorch}.

Using a classifier trained on MNIST, for experiments on this dataset we also plot the \textbf{entropy} of the generated samples' mode distribution, as well as the \textbf{total variation} between the class distribution of the generated samples and a uniform one. For some of the toy experiments, we also used the log-likelihood.

We highlight that since the goal of SGAN is to produce a \textit{single} generator, here we do not compare to ensemble methods. Instead, we demonstrate that samples of the global generator taken at any iteration are of a higher quality, compared to those taken from the local ones.

For further results and details on the implementation, see the Appendix.

\subsection{Experimental results on toy datasets}\label{sec:exp}

Primarily, to motivate the idea of favoring information from the independent ensemble to train a pair, we conduct the following experiment. We train in parallel few pairs of networks, as well as \textit{two} additional pairs: 
\begin{enumerate*}[series = tobecont, itemjoin = \quad, label=(\roman*)]
\item \textbf{SGAN} trained with the local independent pairs, as well as
\item \textbf{GAN} a regularly trained pair.
\end{enumerate*}
In addition to training these two pairs with equal frequency, we used the \textit{identical real-data and noise samples}. 

Figure~\ref{fig-ensemblegan} depicts such an experiment, where we used the vanilla-GAN algorithm and trained on the 10-GMM dataset (Figure~\ref{subfig-real_data}). 
We recall that the only difference between the two discriminators is that the GAN discriminator is trained with fake samples from his tied single opponent (Figure~\ref{subfig-direct_disc}), whereas the one of SGAN is trained with fake samples from the ensemble (Figure~\ref{subfig-indirect_disc}). 

Figure~\ref{subfig-modes_not_cov} depicts that the probability that a mode will not be covered (y-axis) by the ensemble, at a random iteration, goes down exponentially with the number of pairs (x-axis). To this end, we used the 8-GMM toy dataset and vanilla-GAN.

\begin{figure}[ht!]
\includegraphics[width=\linewidth,trim={4.1cm 1cm 3.7cm 2cm}, clip]{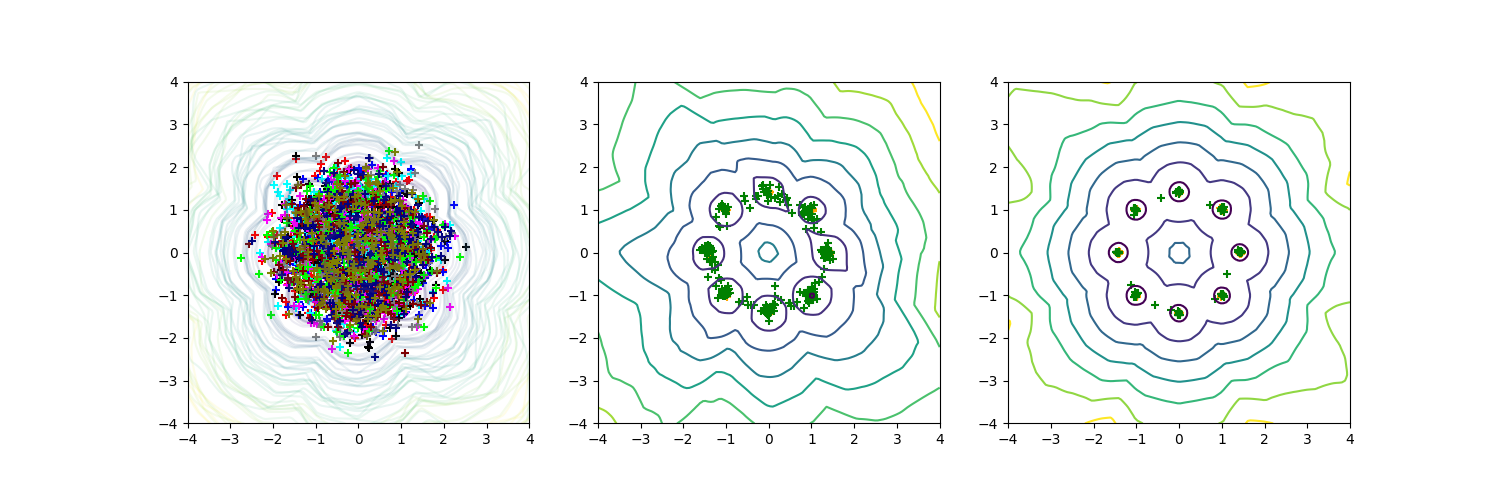}

  \includegraphics[width=\linewidth,trim={4.1cm 1cm 3.7cm 2cm}, clip]{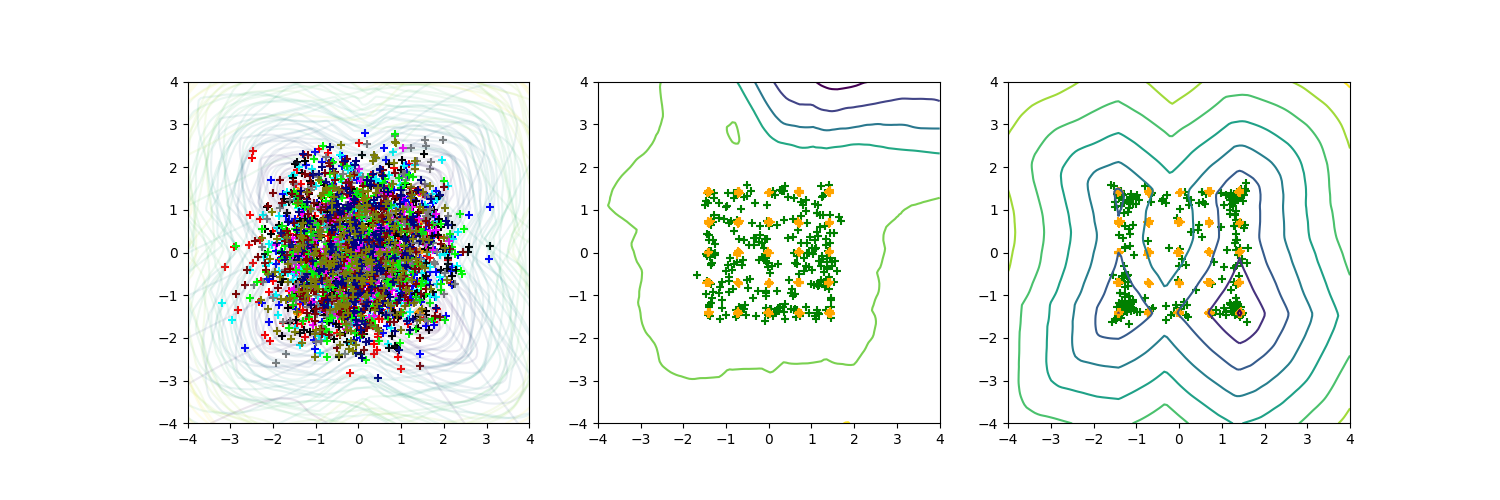}

  \includegraphics[width=\linewidth,trim={4.1cm 1cm 3.7cm 2cm}, clip]{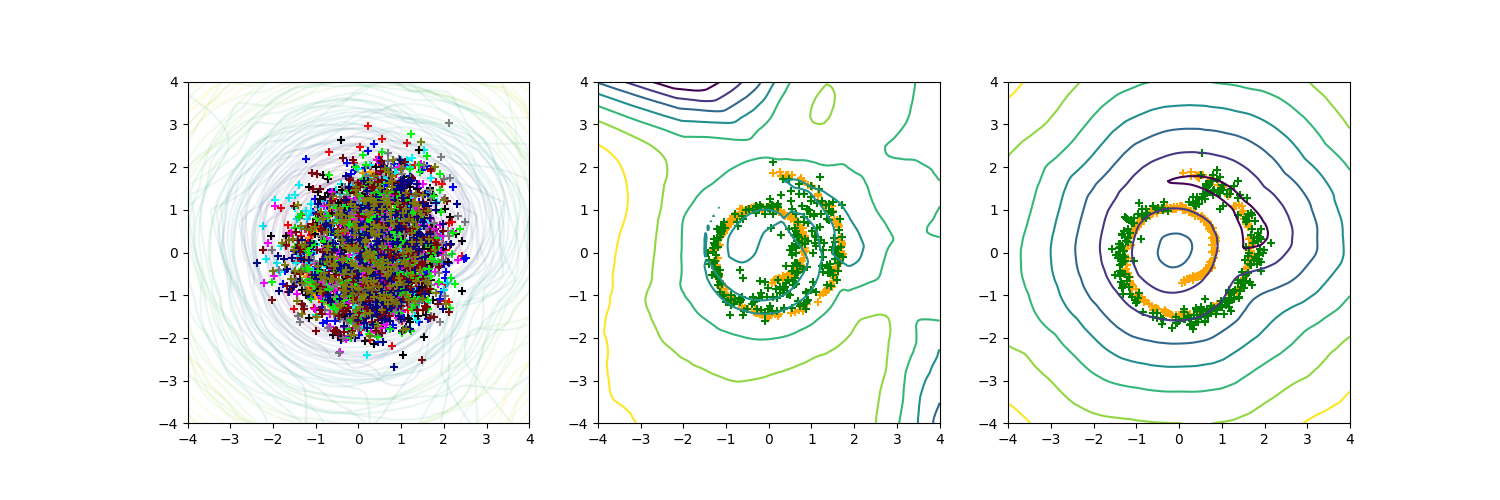}
  \caption{
    (10-S-)WGAN on (top to bottom row): circle 8-GMM, grid 25-GMM, Swiss Roll (best seen in color).
    Real data-points are shown in orange, and yellow and purple contours denote low and high, respectively.
    The first column depicts the $10$ local pairs: generators' samples and discriminators' contours (level sets) are displayed in varying and transparent colors, respectively.
The rightmost column depicts the $10$-S-WGAN  pair (trained with the networks of the first column): samples from $G_0$ are drawn in green, whereas the illustrated contours are from $D_0$.
    The middle column illustrates standard pair trained  $N$-fold more times (see text).
	}
	\label{fig-wgantoy}
\end{figure}

In Figure~\ref{fig-wgantoy} we use WGANs.
We observe that S-WGAN exhibits higher stability and faster convergence speed.
Figure~\ref{fig-wgantoy} also depicts samples from a generator updated $N$-fold times more (middle column), what indicates that a SGAN generator is comparable with these.

\begin{figure}[H]
\centering
\begin{minipage}{\linewidth}
		\includegraphics[width=\linewidth,trim={0cm 1.2cm 0cm 1.85cm}, clip]{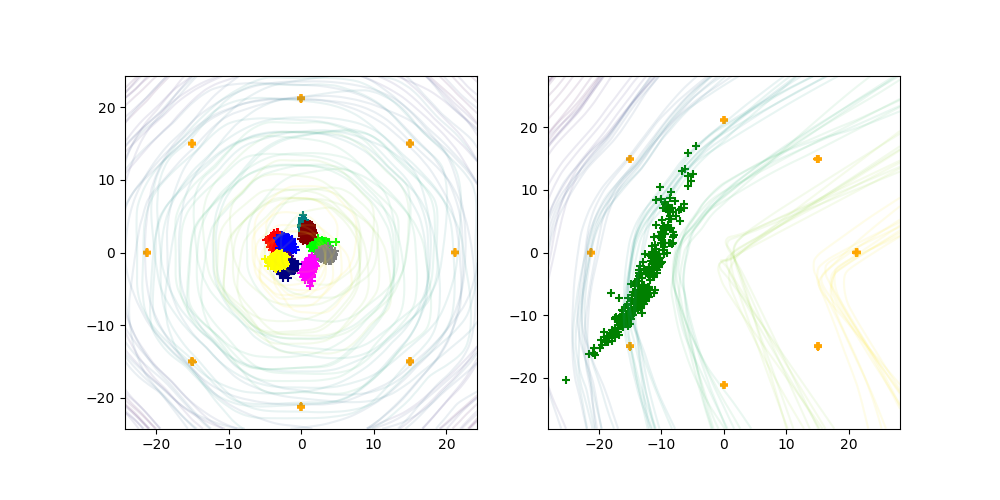}		
	\end{minipage}
\begin{minipage}{\linewidth}
			\includegraphics[width=\linewidth,trim={0cm 1.2cm 0cm 1.5cm}, clip]{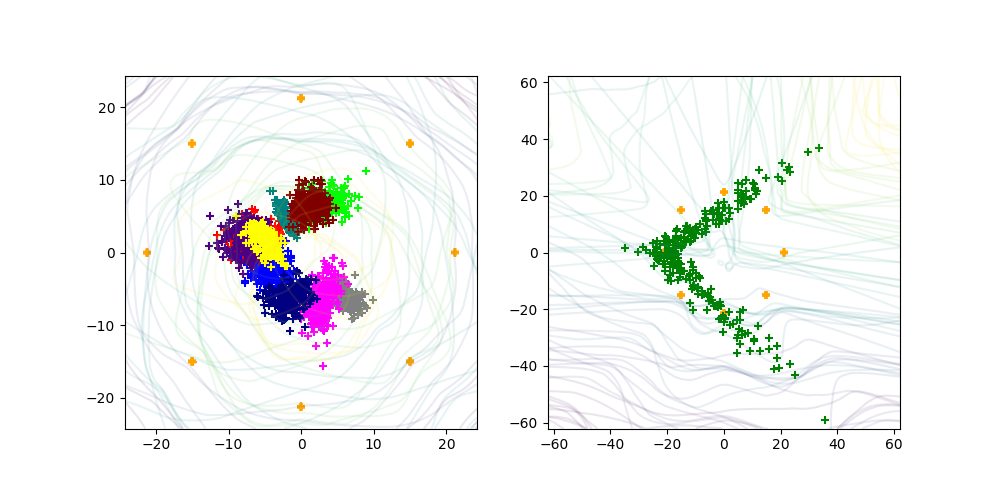}
	\end{minipage}
\begin{minipage}{\linewidth}
		\includegraphics[width=\linewidth,trim={0cm 1.2cm 0cm 1.5cm}, clip]{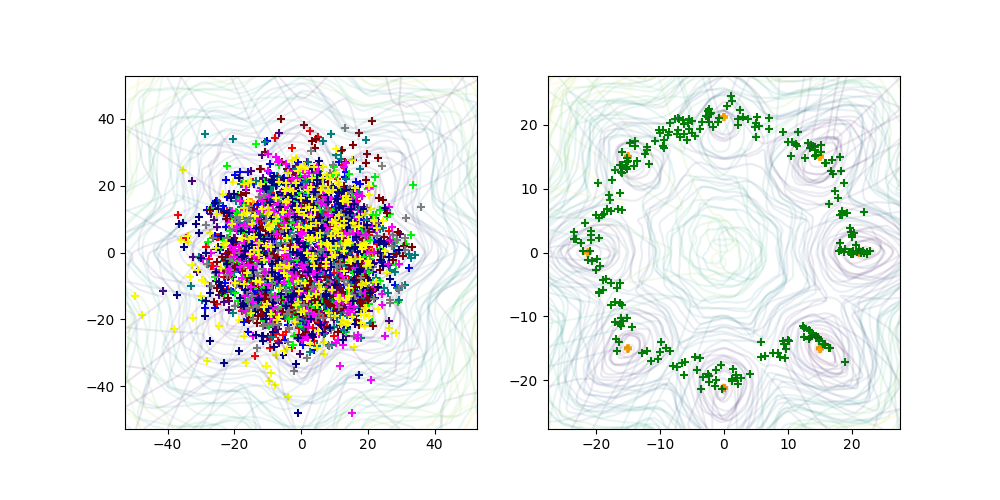}
	\end{minipage}
\begin{minipage}{\linewidth}
		\includegraphics[width=\linewidth,trim={0cm 1.2cm 0cm 1.5cm}, clip]{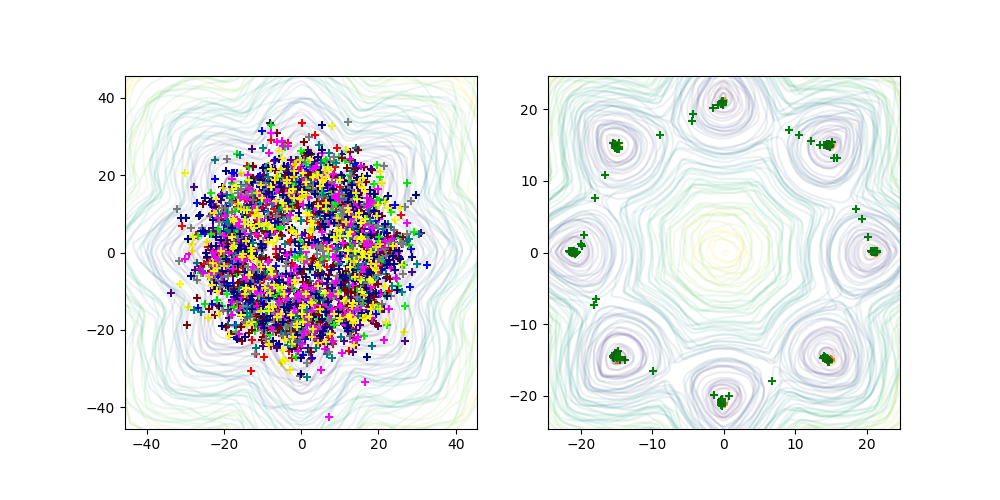}
	\end{minipage}
	\caption{10-S-WGAN on the 8-GMM toy dataset (best seen in color). 
	Samples from $p_d$ are displayed in orange.
	Each row is a particular iteration (top to bottom): 5th, 10th, 100th, and 400th iteration.
	Samples from the local generators and the global one are illustrated on the left (in separate colors) and right (in green), respectively.
	The displayed contours represent the level sets of $\mathcal{D}$ and $\mathcal{D}^{msg}$--illustrated on the left and  right, respectively, where yellow is low and purple is high.
	}
	\label{fig-g0jointLearnp15}
\end{figure}

In Figure~\ref{fig-g0jointLearnp15} we use 10-S-WGAN.
After training the local pairs, the global $G_0$ is trained with the local discriminators, samples of which are displayed on the left and right, respectively.
We observe that at the very first iterations it may be pushed away further than where the real data lies.
Nonetheless, notably, it converges much quicker, as it does not ``explore'' regions already ``visited'' by the local ones.

\subsection{Experimental results on real-world datasets}\label{sec:results-real-data}

\begin{figure*}[!htb]
    \centering
    \begin{subfigure}[t]{0.33\linewidth}
        \centering
        \includegraphics[width=\linewidth]{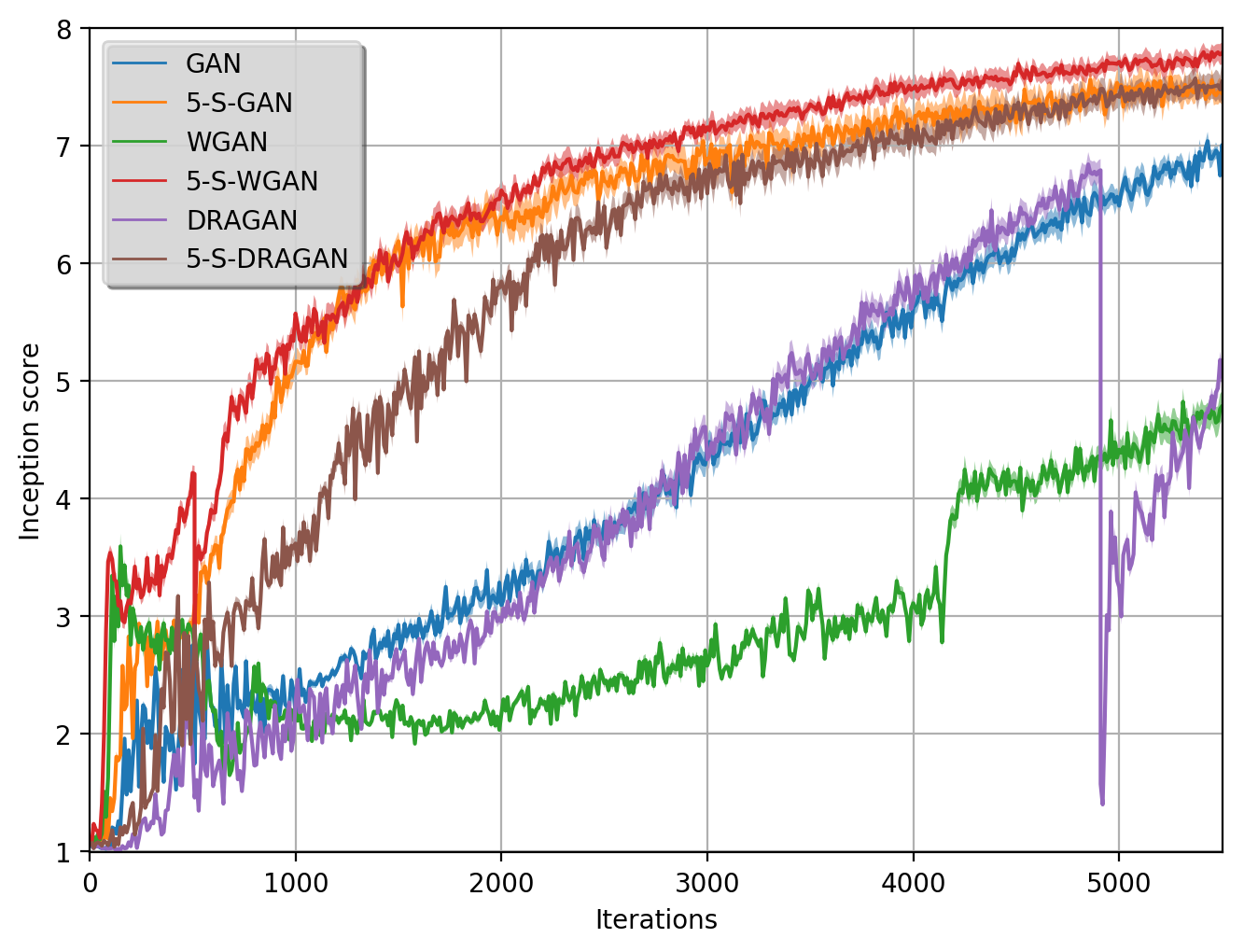}
        \caption{Inception score~\cite{Salimans2016improvingGANs} (higher is better)}\label{subfig-mnist_is}
    \end{subfigure}
    \begin{subfigure}[t]{0.33\linewidth}
        \centering
        \includegraphics[width=\linewidth]{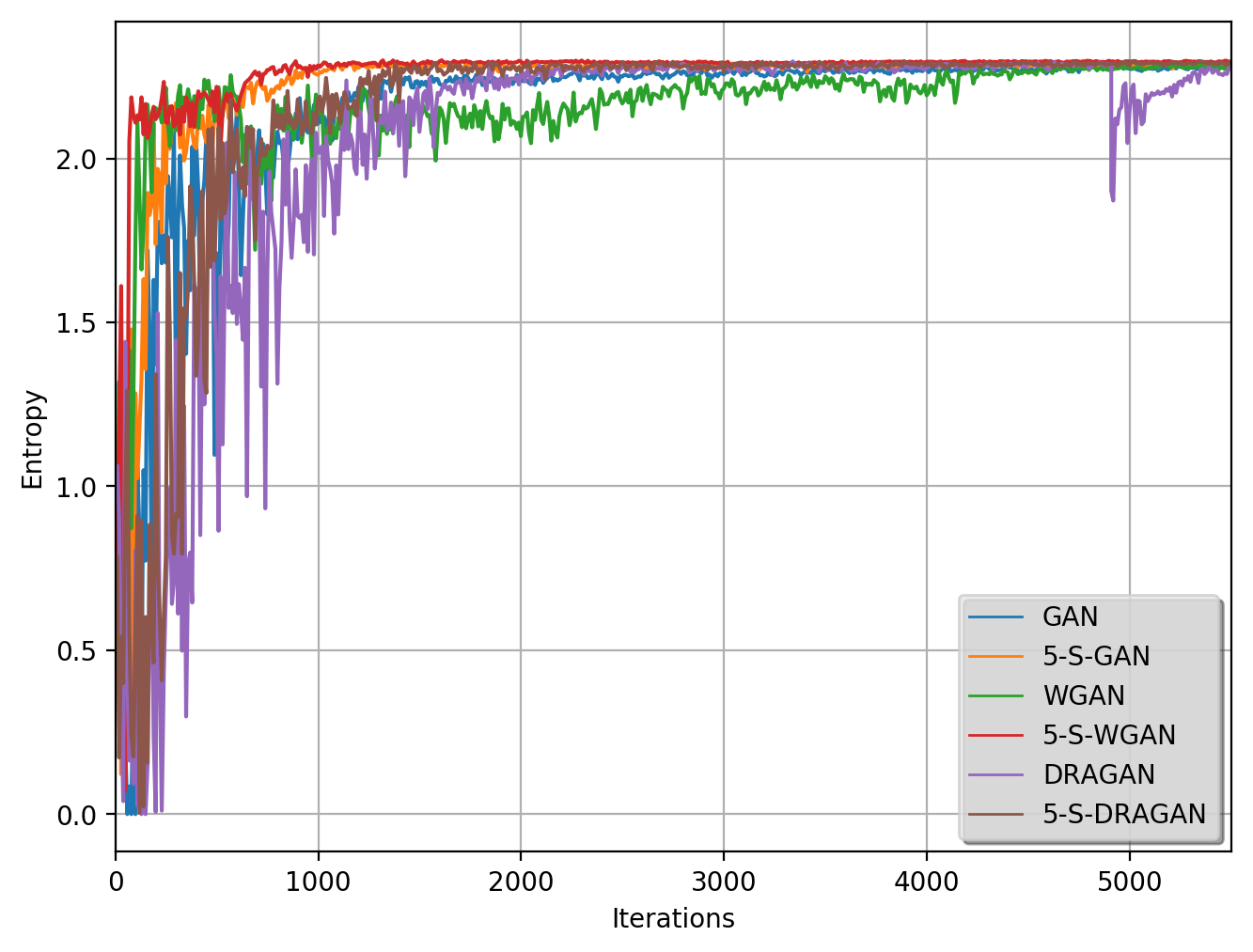}
        \caption{Entropy (higher is better) }\label{subfig-mnist_entropy}
    \end{subfigure}
    \begin{subfigure}[t]{0.33\linewidth}
        \centering
        \includegraphics[width=\linewidth]{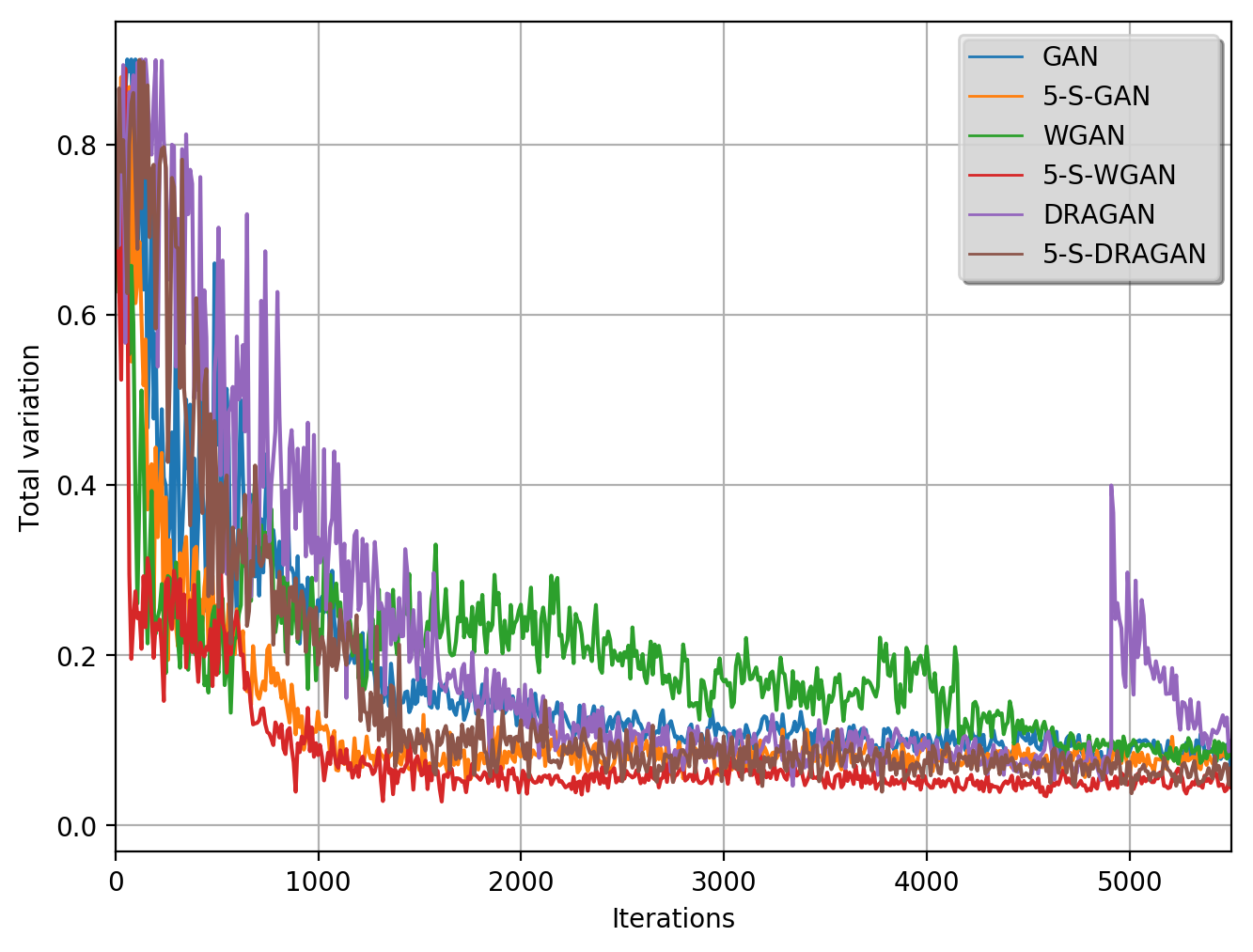} 
        \caption{Total variation (lower is better)}\label{subfig-mnist_tv}
    \end{subfigure}
    \caption{ Results on MNIST using (5-S-) GAN/WGAN/DRAGAN (best seen in color).}
	\label{fig-mnist}
\end{figure*}

\begin{figure}[!htb]
	\noindent
    \begin{subfigure}[t]{0.15\linewidth}
        \begin{tabular}{c}
        \includegraphics[width=\linewidth]{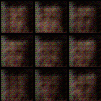} \\
        \includegraphics[width=\linewidth]{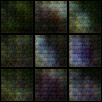}\\
        \includegraphics[width=\linewidth]{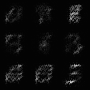} \\
        \includegraphics[width=\linewidth]{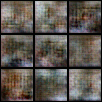}
        \end{tabular}
        \vspace{-.5\baselineskip}
        \caption{$G_1$}\label{subfig-g1}
    \end{subfigure} 
    \begin{subfigure}[t]{0.15\linewidth}
        \begin{tabular}{c}
        \includegraphics[width=\linewidth]{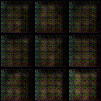} \\
        \includegraphics[width=\linewidth]{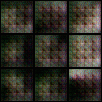}\\
        \includegraphics[width=\linewidth]{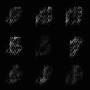}\\
        \includegraphics[width=\linewidth]{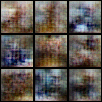}
        \end{tabular}
        \vspace{-.5\baselineskip}
        \caption{$G_2$}\label{subfig-g2}
    \end{subfigure} 
    \begin{subfigure}[t]{0.15\linewidth}
        \begin{tabular}{c}
        \includegraphics[width=\linewidth]{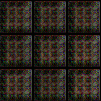} \\
        \includegraphics[width=\linewidth]{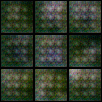}\\
        \includegraphics[width=\linewidth]{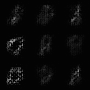}\\
        \includegraphics[width=\linewidth]{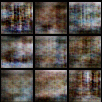}
        \end{tabular}
        \vspace{-.5\baselineskip}
        \caption{$G_3$}\label{subfig-g3}
    \end{subfigure} 
    \begin{subfigure}[t]{0.15\linewidth}
        \begin{tabular}{c}
        \includegraphics[width=\linewidth]{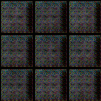} \\
        \includegraphics[width=\linewidth]{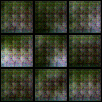}\\
        \includegraphics[width=\linewidth]{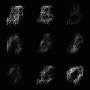}\\
        \includegraphics[width=\linewidth]{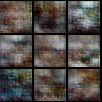}
        \end{tabular}
        \vspace{-.5\baselineskip}
        \caption{$G_4$}\label{subfig-g4}
    \end{subfigure} 
    \begin{subfigure}[t]{0.15\linewidth}
        \begin{tabular}{c}
        \includegraphics[width=\linewidth]{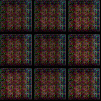} \\
        \includegraphics[width=\linewidth]{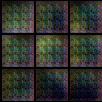}\\
        \includegraphics[width=\linewidth]{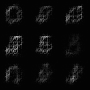}\\
        \includegraphics[width=\linewidth]{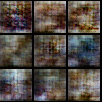}
        \end{tabular}
        \vspace{-.5\baselineskip}
        \caption{$G_5$}\label{subfig-g5}
    \end{subfigure} 
    \begin{subfigure}[t]{0.15\linewidth}
        \begin{tabular}{c}
        \includegraphics[width=\linewidth]{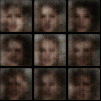} \\
        \includegraphics[width=\linewidth]{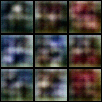}\\
        \includegraphics[width=\linewidth]{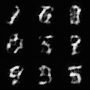}\\
        \includegraphics[width=\linewidth]{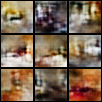}
        \end{tabular}
        \vspace{-.5\baselineskip}
        \caption{$G_0$}\label{subfig-g0}
    \end{subfigure} 
    \caption{\textbf{5-SGAN}. In the top to bottom rows we use DCGAN on CelebA, DRAGAN on ImageNet, WGAN on MNIST and DCGAN on LSUN, respectively. Each of the above samples are taken at the earlier iterations, in particular at the 100\textit{-th}, 500\textit{-th}, 500\textit{-th} and the 1000\textit{-th} iteration, respectively for each row. In columns (\subref{subfig-g1}-\subref{subfig-g5}) we show samples from the local generators, whereas  in (\subref{subfig-g0}) from the global generator. We used separate networks, and real data space of $32{\times}32$. }
    \label{fig-imgdatasets}
\end{figure}

In Figures~\ref{fig-mnist} and \ref{fig-imgdatasets} we show experimental results on image datasets. In the latter, samples are taken at a random iteration, \textbf{prior to final convergence}, so as the difference in the quality of the samples is clearer. In Table~\ref{cifar10_inception} we list quantitative results on CIFAR10, using the Inception score~\cite{Salimans2016improvingGANs}. The architectures are held fixed for all the listed experiments in Table~\ref{cifar10_inception}. All of the SGAN implementations are separate networks, that is we do not use weight sharing, and the regular training of one pair is always a separate experiment, rather than getting the scores of the local pairs--to ensure proper sampling. The hyperparameters of the one pair training of DCGAN in Table~\ref{cifar10_inception} have been tuned and \textit{we list the scores of the best performing experiment}, whereas for \textit{all} of the rest of the experiments we use a default fixed setup.

In Table~\ref{tab:billionWords} we show snippets of the fake samples when training SGAN on the One Billion Word Benchmark dataset.  It is interesting to observe similar behavior as on toy datasets: at the first iteration (first row in Table~\ref{tab:billionWords}) the SGAN generators are pushed far from the modes of the real data samples, as they generate non-commonly used letters.  However, SGAN generators converge faster compared to a standard generator 
as they quickly start to generate most commonly used characters.

In summary, the results indicate that SGAN converges faster and has a better coverage of the density. When generating images in particular, it more rapidly and accurately models the local statistics and fixes the grainy texture visible on samples generated by the baseline.

\begin{table}
\caption{ Inception scores~\cite{Salimans2016improvingGANs} on CIFAR10.
From top to bottom rows, we list the obtained scores at the 100\textit{-th}, 500\textit{-th}, 1000\textit{-th}, 2000\textit{-th}, 3000\textit{-th}, 4000\textit{-th} and the 5000\textit{-th} iteration. See text.}\label{cifar10_inception}
\begin{adjustbox}{max width=\linewidth}
\begin{tabular}{c@{\hskip 0.09in}c@{\hskip 0.09in}c@{\hskip 0.09in}c@{\hskip 0.09in}c@{\hskip 0.09in}c}
 WGAN      & 5-S-WGAN   	&  DRA    &  5-S-DRA   & GAN 		& 5-S-GAN \\
\hline
1.05(.000) & 1.45(.004) &  1.06(.001) &  1.16(.002)& 1.04(.000) & 1.37(.004)\\
1.12(.001) & 1.67(.007) &  1.44(.006) &  1.58(.005)& 1.63(.004)	& 1.62(.013)\\
1.14(.002) & 2.25(.015) &  1.32(.004) &  2.20(.011)& 1.71(.014)	& 2.45(.021)\\
1.19(.001) & 3.06(.032) &  2.07(.011) &  3.36(.039)& 2.51(.013)	& 3.41(.033)\\
1.19(.002) & 3.81(.037) &  2.38(.017) &  3.73(.064)& 3.25(.029)	& 4.04(.046)\\
1.19(.002) & 3.99(.036) &  3.05(.022) &  4.17(.051)& 3.67(.037)	& 4.38(.039)\\
1.21(.002) & 4.44(.058) &  3.40(.036) &  4.80(.069)& 3.58(.020)	& 4.73(.051)\\
\hline
\end{tabular}
\end{adjustbox}
\end{table}

\begin{table}[!ht]

\centering
\caption{Output snippets of the global generators trained on the One Billion Word
  Benchmark. In each row we show samples taken at a 
  particular iteration, where top to bottom row: 1\textit{-st}, 100\textit{-th}, 200\textit{-th} iteration.  
  The output of a standard WGAN training is \textbf{a single character (white space) 
  for all of the first 660 iterations}, what indicates slower then $N$-fold convergence speed compared to a N-S-Generator.
}
\label{tab:billionWords}
\begin{adjustbox}{max width=\linewidth}
\begin{tabular}{cc}
 5-S-WGAN & 10-S-WGAN \\
	\begin{minipage}{.48\linewidth}
	\centering
\begin{Verbatim}[fontsize=\tiny, frame=single]
粪ğÄуqуÇ粪粪\{€⁈Äуm&»♠у→№粪у粪у粪&u㭡㭡Ň
Äуß8&İуууóууğuğqAğÄ{?M㭡粪ü8уу8粪уq
ü粪q粪q粪M{уÒşğ½усğуя粪İ&ą8оо粪粪Y粪ÄЄу
'уİûτ㭡уAŞ&Hğ{m7(ğÃHqуŇİ粪уµH^H粪ğ{
уóу{у&Ä{粪ŇÍуﬁ˚İq{∆ş€уğq0&ğiðğ粪`v
у♠qу粪粪у粪у&uą&粪粪оóMšÒάŇя8m∆ğ粪?粪Çğ
İуmуŸÄу粪8α粪оğ&у粪АtHqq9{粪ımуv粪→㭡8
óуżłуAHуÍğİ粪ğÄİHÇуğуŇ﻿ąİ&q№sİ»уı
уоğÄуŇ粪у粪粪qу粪Ňİ粪q!m{ğğŇу♠Äİ¾€ш→粪
\end{Verbatim}
\end{minipage}  &
	\begin{minipage}{.48\linewidth}
	\centering
\begin{Verbatim}[fontsize=\tiny, frame=single]
m`¨mam€ˆúåˇHÄń¦Ňń¨åÄ»ŇÄÈt»Ãń
Äå¾Äˆ`¨ÄâÇÄŇˇ¨¾1å(MMńÄńк
ÄÄˇåÇAåÄ¨ÈÄ8qqÄÈÈqθÈ•HÄ&åÄÄÄ
ÄÄθÄ¨óHńП&ÄńńÄwŵóÈ»€♪¦ÄÄÈÄw©ńθˇ
ÄˇŇéAsˇÄöÄÄsŇÄŇ¾ÈÍÄ8qÊmńsqvńÄ
Í¨¨​Ä•¦mзŇŵěA¨ÄńƒÈÄsˇmmˇˇ`
θYˇ1ÄÄÄÍ&s€&ˇ1Á¨ń​‟ěÈˇÄu•8Èå
»ÄÈmń¨Ä¨Ä♪Ä→åW6]ńÄ9Äm&ńÄÄÄâ?ó
ûÄAâˇθuWÄÇÍH´1Ä&Äńịm􀀗&ÄÄˇ
\end{Verbatim}
\end{minipage} \\
\begin{minipage}{.48\linewidth}
	\centering
\begin{Verbatim}[fontsize=\tiny, frame=single]
aaa aaaaaaaaaaaaaa aaaa aaaaaaaa
a   aaaaaa aaaaaaaa aaaaa aaaaaa
 aaaaaa a aaaaaaaaa aaaaa aaaaaa
 aaa aaaaaaaa aaaaaaaaaaaaaaaa a
a aaaaaaaaaaa aaaaaaaaa aaaaaaaa
  a aaaa aaa aaaaaa aaaaaaaaa a
  aaaa  aaaaaaaaaaaaa aaa aaaa a
aaaaaa  aaaa a aaaaaaaaaa aaaaaa
a  aaa  aa aaaa aaaaaaaaaaaaaaa
\end{Verbatim}
\end{minipage}  &
	\begin{minipage}{.48\linewidth}
	\centering
\begin{Verbatim}[fontsize=\tiny, frame=single]
ii iii   iiiii   iiiii  i iiiii
ii  ii iiiii ii iiii i iiiii ii
iiiii ii ii iii ii iiii i   iiii
iiii iii i  i  iiii iiiiii  i
iii iiiii   i  iii i i ii i iii
ii ii     iii i iii ii ii iii ia
iiii  iii iiiiii  i  i   i i  ii
i iiii iiiii i ii iiiii iii   i
iiiiii   i i  iiiii     i i i ii
\end{Verbatim}
\end{minipage} \\
\begin{minipage}{.48\linewidth}
	\centering
\begin{Verbatim}[fontsize=\tiny, frame=single]
hieq  as ieq aa  dhhie  as ie  t
eq shiq as heq aaa hheq  asheq t
ieqq aasheq dsheq aa dd dhhie  t
iq ddq as ie  sie  ashieq as e t
eq as hheq  aas ie  s heq as e t
q hheq aas ieq dshieq as hie  at
hhhieq aa  ieq asshiq  aas ieq t
qq hieq as ie  dsieq as hhhie  t
eq asid ddd as hieq as heq diq t
\end{Verbatim}
\end{minipage}  &
	\begin{minipage}{.48\linewidth}
	\centering
\begin{Verbatim}[fontsize=\tiny, frame=single]
SAeS Aer areS SnSSSharSonS Soe
AS SSSSer oeS SarSonSSS Ss ShS i
S tes SrSsne SoerhaS SsnS Soar o
MSSSS S Sha tos ShS aoS as Sha i
s She tAeS SsS SsnSSSoes ShS Son
Aes s S iSnSShar SoSe  anSSsnS S
eS es S SoS Ss SoSSS Ss SSS Son
ASS SeS as SnSrSar SsrSrer SnS s
ASSSS tSa SoarSonS Ssne Soar osn
\end{Verbatim}
\end{minipage} \\

\end{tabular}
\end{adjustbox}
\end{table}

\section{Related work: Multi-network GAN methods}\label{sec:related-multi-network}
Independently, Boosted Generative Models~\cite{boostedGAN} and AdaGAN~\cite{2017arXivAdaGAN} propose the iterative boosting algorithm to solve the mode collapse problem. At each step a new component is added into a mixture of models, by updating the samples' weights, while using the vanilla GAN algorithm. 

With similar motivation of increasing the mode coverage ~\cite{multiagent} and ~\cite{2017arXiv170802556H} propose to instead train multiple generators versus a single discriminator.

In~\cite{multiagent} the discriminator is trained against $N$ generators which share parameters in all layers except the last one, and it outputs probability estimate for $N{+}1$ classes representing whether the input is a real sample, or by whom of the generators it originates. To enforce diversity between the generated samples, a penalty term is added with a user-defined similarity based function.

Similarly,~\cite{2017arXiv170802556H} proposes multiple generators that share parameters versus single discriminator whose output is fake versus real, as well as training an additional model that classifies by whom of the generators a given fake input was generated. The output of the classifier is used in an additional penalty term that forces diversity between the generators.
\cite{DurugkarGM16}  proposes utilizing multiple discriminators versus one generator, in aim to stabilize the training.

\cite{GhoshKN16}  proposes multiple generators versus single discriminator, where the generators communicate through two types of messages. Namely there are co-operation and competing objectives.
The former ensures the other generator to generate images better then itself, and the latter encourages each generator to generate better samples then its counterpart.

Motivated by the observed oscillations in \cite{ensembleGAN} a so called ``self-ensembles'' is proposed. Non-traditionally, this self-ensemble is built out of copies of the generator 
taken from a different iteration while training a single pair.

Hence, SGAN depicts different structures and solutions to the problem of training GANs.
Regarding the former, none of the above methods utilizes explicitly multiple pairs trained independently. Instead, most commonly a structure of one-to-many is used, either for the generator or for the discriminator. Compared to AdaGAN, SGAN is applicable to any GAN variant,  runs in parallel, and produces a single generator.
Concerning the latter, SGAN uses ``supervising'' models and prevents an influence of one pair towards all.

\section{Different viewpoints of SGAN}\label{ss-SGAN_in_game_theory}
\paragraph{Connecting SGAN to Actor-critic methods.}

In~\cite{PfauV16} the authors argue that at an abstract level GANs find similarities with actor-critic (AC) methods, which are widely used in reinforcement learning.
Namely, the two have a feed-forward model which either takes an action (AC) or generates a sample (GAN).
This acting/generating model is trained using a second one.
The latter model is the only one that has direct access to information from the environment (AC) or the real data (GAN), whereas the former has to learn based on the signals from the latter. We refer the interested reader to~\cite{PfauV16} which further elaborates the differences and finds connections that both the methods encounter difficulties in training.

\begin{figure}[!htb]
    \centering
    \begin{subfigure}[t]{.45\linewidth}
        \centering
        \includegraphics[width=\linewidth,trim={0cm 0cm 0cm 0cm}, clip]{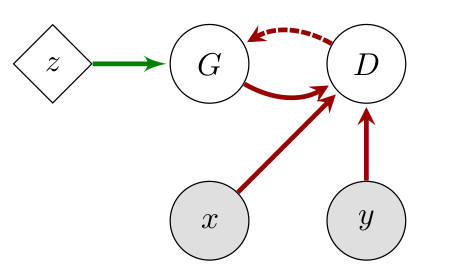}
        \caption{GAN training~\cite{PfauV16}}\label{subfig-ac_gan}
    \end{subfigure} 
    \begin{subfigure}[t]{.45\linewidth}
        \centering
        \includegraphics[width=\linewidth]{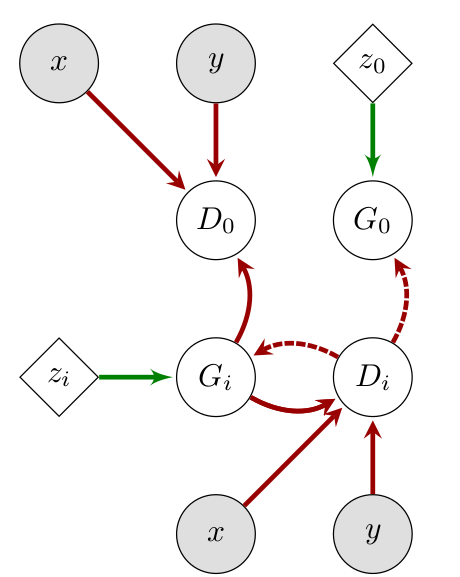}
        \caption{SGAN training}\label{subfig-ac_sgan}
    \end{subfigure}
    \caption{	Graphical representations~\cite{PfauV16} of the information flow structures of GAN and SGAN training. Subfigure (\subref{subfig-ac_gan}) depicts a connection between GAN  and Actor-critic methods, proposed in~\cite{PfauV16}. In (\subref{subfig-ac_sgan}) we extend the former, to illustrate the case of SGAN training, where nodes with index $i$ can be multiple.
     Empty circles represent models with a distinct loss function. Filled circles represent information from the environment. Diamonds represent fixed functions, both deterministic and stochastic. Solid lines represent the flow of information, while dotted lines represent the flow of gradients used by another model.  }
	\label{fig-rl}
\end{figure}

We make use of the graphical illustration proposed in~\cite{PfauV16} of the structre of the GAN algorithm illustrated in Figure~\ref{subfig-ac_gan}, and we extend it to illustrate how SGAN works, Figure~\ref{subfig-ac_sgan}.
In SGAN, $D_0$ is being trained with samples from the multiple generators whose input is in the real-data space.
For clarity, we omited $D^{msg}$ in the illustration--used to train $G_0$, as the arrows already indicate that these two ``global'' models do not affect the ensemble.

\paragraph{Game theoretic interpretation.}

We can define a game that describes the training of $G_0$ and $D_0$ in the SGAN framework as follows. Let us consider a tuple $(\mathcal{P}, \mathcal{A}, u)$, where $\mathcal{P} = \{G, D\}$ is the set of new players that we introduce.
Let us assume that $G$ and $D$, at each iteration can select among the elements of $\mathcal{D}$ and $\mathcal{G}$, respectively.
Hence, $\mathcal{A} = (A_{g}, A_{d})$ have a finite set of $N$ actions.

Such ``top level players'' in SGAN assign uniform distribution over their actions, more precisely both $G$ and $D$ sample from the elements of $\mathcal{D}$  and $\mathcal{G}$ respectively, with uniform probability.
To connect to classical training, let us assume that $G$ and $D$ fix their choice to one  element of $\mathcal{D}$ and $\mathcal{G}$ respectively, \latin{i.e.} with probability one they sample from a single generator/discriminator.
The trained networks $G_0$ and $G_i$, as well as $D_0$ and $D_j$, with $i$ and $j$ being the selected choice of $G$ and $D$ respectively, are identical in expectation.
Finally, rather than predefining the uniform sampling in SGAN, incorporating estimations of the actions' pay-off $u=(u_g, u_d)$ could prove useful for training $(G_0, D_0)$.

\section{Conclusion}\label{sec-conclusion}

We proposed a general framework dubbed SGAN for training GANs, applicable to any variant of this algorithm. It consists of training several adversarial pairs of networks independently and uses them to train a global pair that combines the multiple learned representations.

Motivated by the practical difficulties of the training, SGAN builds upon a straightforward idea, and yet it proves itself as a very powerful framework. To our knowledge, it is the first method that directly addresses the discrepancy between the theoretical justifications being derived in functional space, and the fact that we optimise the parameters of the deep neural networks~\cite{GoodfellowGAN2014}.

A key idea in our approach is maintaining the statistical independence between the individual pairs, by preventing any flow of information between them, in particular through the global pairs it aims at training eventually. Maintaining this makes the probability of a failure to go down exponentially with the number of pairs involved in the process.

An experimental validation on very diverse datasets demonstrates that this approach systematically improves upon classical algorithms and that it provides a much more stable framework for real-world applications. Furthermore, SGAN is convenient for many applications in computer vision, as it produces a single generator.

Some future extensions of SGAN include improving the covering behavior by forcing diversity between the local pairs and re-casting the analysis in the context of multi-player game theory.

\section*{Acknowledgment}
\noindent
This work was supported by the Swiss National Science Foundation, under the grant CRSII2-147693 "WILDTRACK". We also gratefully acknowledge NVIDIA's support through their academic GPU grant program.

\vspace{1em}
{\small
\bibliographystyle{ieee}
\bibliography{sgan}
}

\clearpage
\appendix
\section{Experiments on toy datasets (an extension)}\label{app:toyexp}
\begin{figure}[!htb]
\centering
\begin{tabular}{c}
	\begin{minipage}{\linewidth}
		\begin{tabular}{c@{\hskip 0in}c}
		\includegraphics[width=.49\linewidth,trim={1cm .7cm 1cm 0.5cm}, clip]{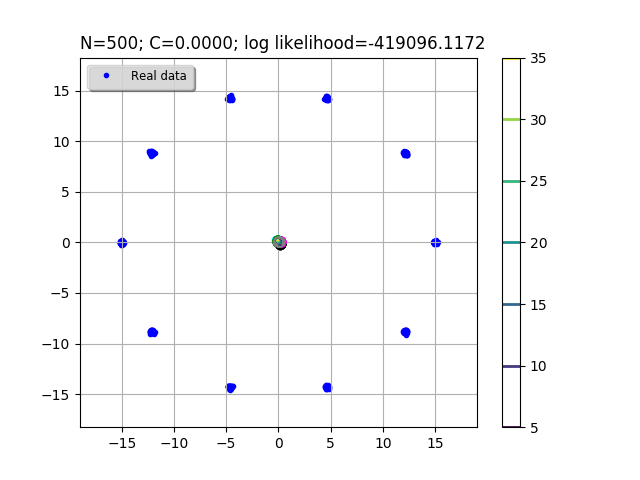} &
		\includegraphics[width=.49\linewidth,trim={1cm .7cm 1cm 0.5cm}, clip]{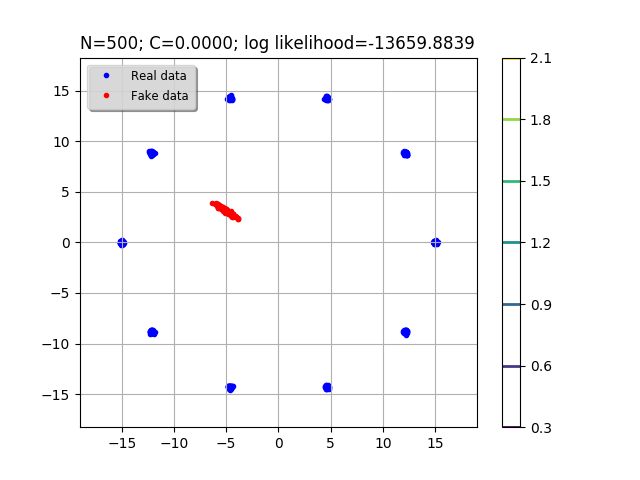}	\\
		\multicolumn{2}{c}{\textit{\small Iteration 1}}  \vspace{.7em}	
		\end{tabular}	
	\end{minipage} 		\\ 
	
	\begin{minipage}{\linewidth}
		\begin{tabular}{c@{\hskip 0in}c}
		\includegraphics[width=.49\linewidth,trim={1cm .7cm 1cm 0.5cm}, clip]{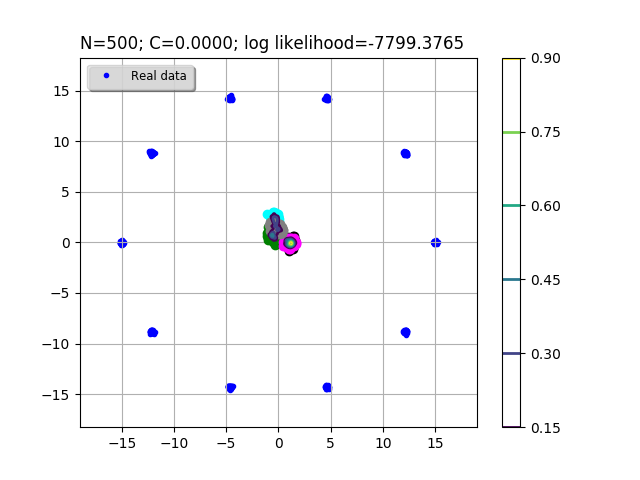} &
		\includegraphics[width=.49\linewidth,trim={1cm .7cm 1cm 0.5cm}, clip]{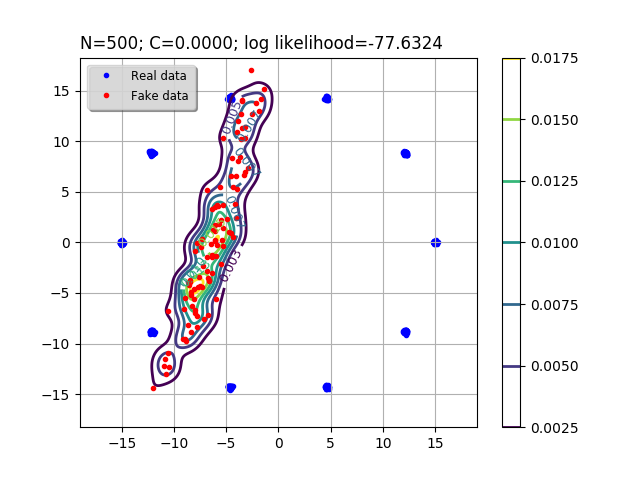}	\\
		\multicolumn{2}{c}{\textit{\small Iteration 8}}	\vspace{.7em}\\
		\end{tabular}	
	\end{minipage} 	 \\
	
	\begin{minipage}{\linewidth}
		\begin{tabular}{c@{\hskip 0in}c}
		\includegraphics[width=.49\linewidth,trim={1cm .7cm 1cm 0.5cm}, clip]{./pics_toy_eglearn_local-e00008.png} &
		\includegraphics[width=.49\linewidth,trim={1cm .7cm 1cm 0.5cm}, clip]{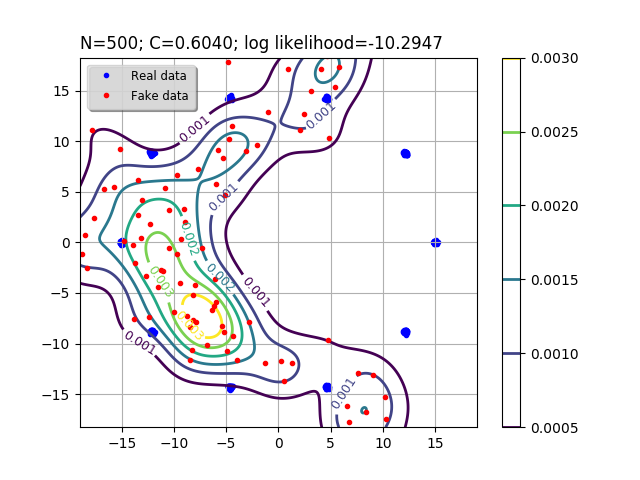}	\\
		\multicolumn{2}{c}{\textit{\small Iteration 40}}	\vspace{.7em}	\\
		\end{tabular}	
	\end{minipage} 		\\ 
	
	\begin{minipage}{\linewidth}
		\begin{tabular}{c@{\hskip 0in}c}
		\includegraphics[width=.49\linewidth,trim={1cm .7cm 1cm 0.5cm}, clip]{./pics_toy_eglearn_local-e00008.png} &
		\includegraphics[width=.49\linewidth,trim={1cm .7cm 1cm 0.5cm}, clip]{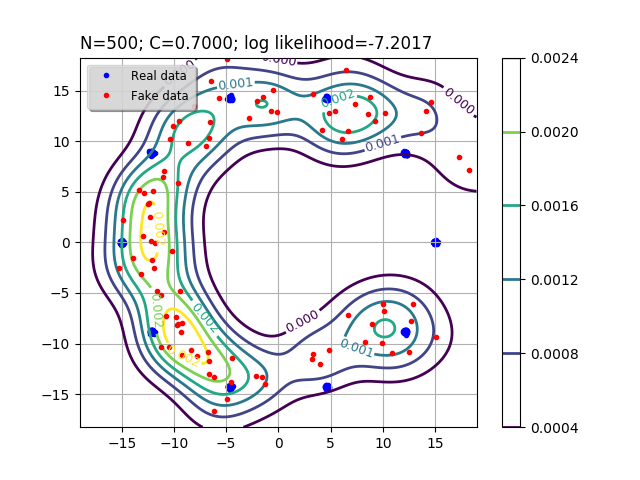}	\\
		\multicolumn{2}{c}{\textit{\small Iteration 125}}	\vspace{.7em}	\\
		\end{tabular}	
	\end{minipage} 	 \\ 
\end{tabular}
	\caption{\textbf{5-S-WGAN} on the \textbf{10-GMM} toy dataset.
	In each image-pair: we illustrate samples from the five local generators and from the global generator, on the left (in separate color) and on the right (in red).
	See text for details, \S~\ref{sec:exp_toy}.
	}
	\label{fig-g0jointLearn}
\end{figure}

\subsection{Details on the implementation}\label{sec:impl_toy}
For experiments conducted on toy datasets we used separate $2{\cdot}(N{+}1)$ networks.
The architecture and the hyper-parameters are as follows. Each network is a multilayer perceptron (MLP) of $4$ fully connected layers and LeakyReLU  non-linearity~\cite{leakyrelu} with the PyTorch's default value for the negative slope of $0.01$~\cite{pytorch}.

The number of hidden units for each of the layers is $512$.
We use a noise vector of a dimension $100$. 
We use learning rate of $1{\cdot}10^{-5}$. 
In most of our experiments we used the \textit{Adam} optimization method~\cite{adam}. We also run experiments with \textit{RMSProp}~\cite{rmsprop} as recommended by the authors of the corresponding GAN variant.
As we observed no obvious improvement favoring one of the above optimization methods on toy datasets, at some point we fixed the choice of the optimization method to \textit{Adam}.

\subsection{Experiments}\label{sec:exp_toy}

\begin{figure}[!ht]
\centering
	\begin{minipage}{.49\linewidth}
		\includegraphics[width=\linewidth,trim={1cm 1.29cm 4cm 1.4cm}, clip]{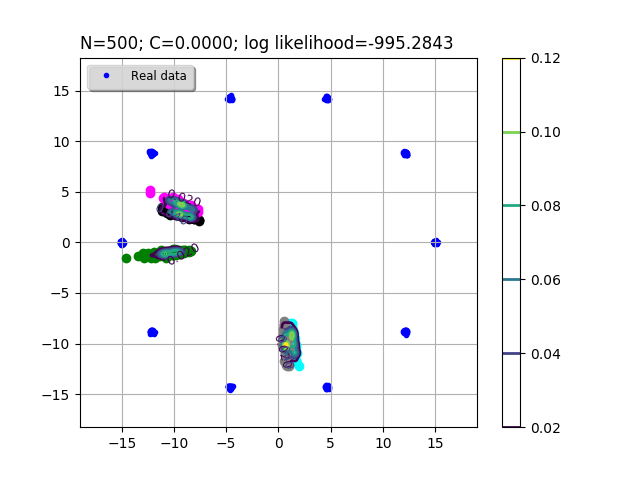}
	\end{minipage}
\begin{minipage}{.49\linewidth}
		\includegraphics[width=\linewidth,trim={1cm 1.29cm 4cm 1.4cm}, clip]{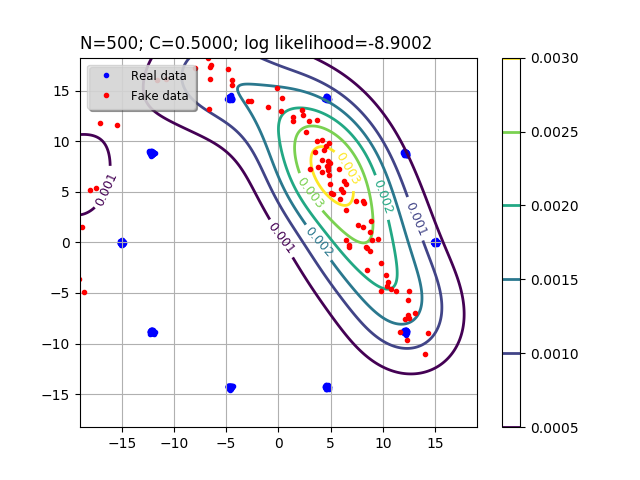}
	\end{minipage}
	\caption{Experiment on the \textbf{10-GMM} toy dataset using a variant of WGAN  (see text) and five local pairs (best seen in color). Real samples, fake samples from the local generators, and fake samples from the global one are illustrated in blue, varying colors on the left, and red color on the right, respectively. 
	The displayed contours are obtained using GMM-KDE with cross-validated bandwidth and $500$ samples of $p_g$.
	}
	\label{fig-g0jointLearn2}
\end{figure}

\begin{figure}[!htb]
\centering
\includegraphics[width=\linewidth]{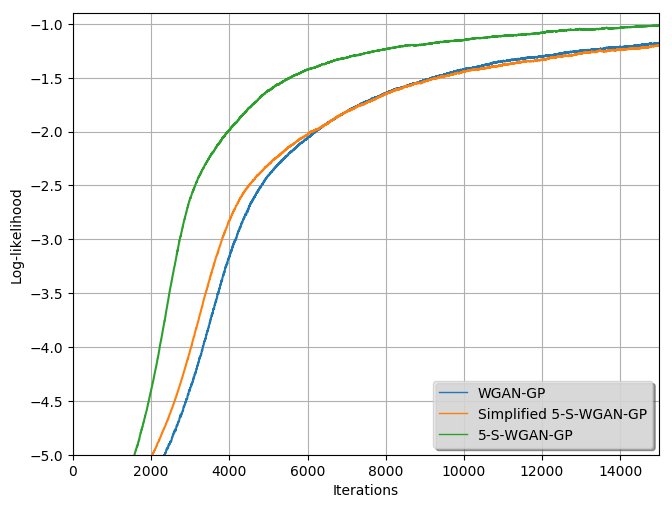}
\caption{
Log-likelihood on \textbf{8-GMM} toy dataset (\S~\ref{sec:exp_toy}).}
\label{fig:log-like-wgangp}
\end{figure}

In the sequel, we use the notation of the methods and the datasets as introduced in \S~\ref{sec-experiments}.

In 	Figure~\ref{fig-g0jointLearn} we illustrate image pairs, where:
\begin{enumerate*}[series = tobecont, itemjoin = \quad, label=(\roman*)]
\item on the left we display samples taken from the local generators; and
\item on the right samples from the global generator.
\end{enumerate*}
To obtain the illustrated contours, we use a GMM Kernel Density Estimation (KDE)~\cite{rosenblatt1956}, with cross-validated bandwidth, and a sample of $p_g$ of size $500$ (in the figure denoted with $N$).
We also implemented the \textit{Coverage} metric, proposed in ~\cite{2017arXivAdaGAN} for toy experiments with GMM (denoted as \textit{C} in Figure~\ref{fig-g0jointLearn}).

As can be observed in Figure~\ref{fig-g0jointLearn}, in the early iterations, samples from the global generator are very likely to be in different regions than those taken from the local generators.
In Figure~\ref{fig-g0jointLearn2} we show similar observations on a different experiment. In the illustrated experiment we used WGAN with gradient penalty, while forcing this constraint in local regions around real samples (as the penalty term in DRAGAN).
In addition, note that samples from $G_0$ may be pushed further from the real data modes in these early stages (for \latin{e.g.} sthe samples on the left of Figure~\ref{fig-g0jointLearn2}).

In Figure~\ref{fig:log-like-wgangp} we plot the log-likelihood on the \textbf{8-GMM} toy dataset.
\textbf{Simplified-5-S-WGAN} denotes the SGAN method without the messengers discriminators (see \S~\ref{sec-method}).

\begin{figure*}[!htbp]
\begin{tabular}{c|@{\hskip 0in}c}
		\begin{minipage}{0.49\linewidth}
		\vspace{1mm}
		\centering
		\includegraphics[width=\linewidth,trim={1.5cm 1cm 1.5cm 1cm}, clip]{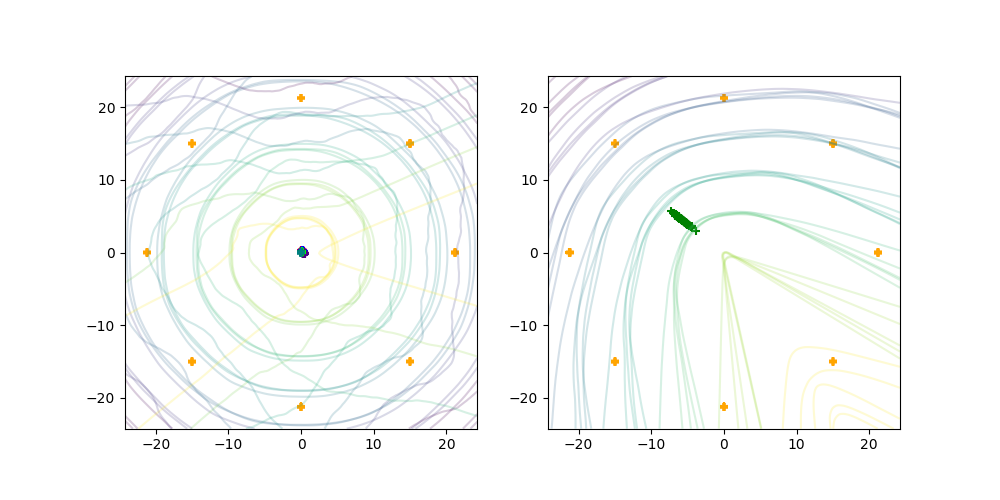}  
		\textit{\small Iteration 1}	 \vspace{1em}
		\end{minipage} & 

		\begin{minipage}{0.49\linewidth}
		\vspace{1mm}
		\centering
		\includegraphics[width=\linewidth,trim={1.5cm 1cm 1.5cm 1cm}, clip]{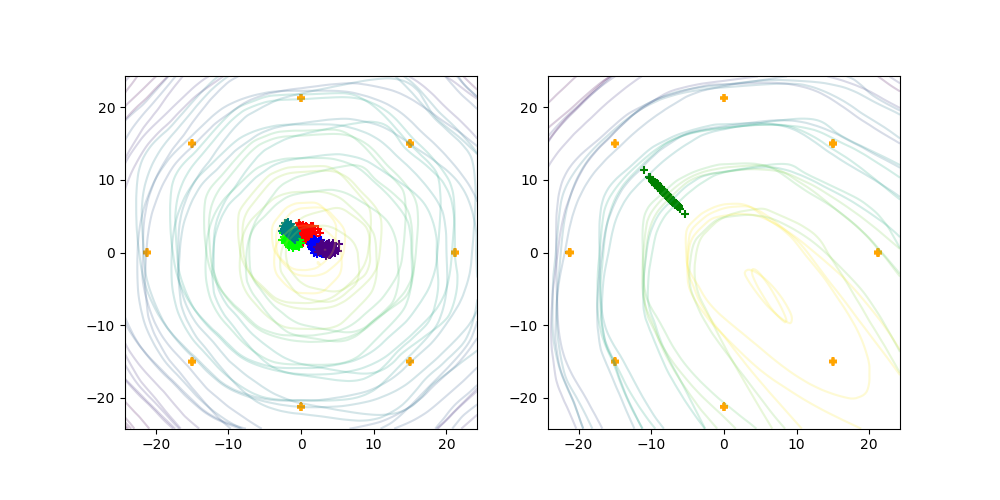}
		\textit{\small Iteration 5}		\vspace{1em}
		\end{minipage} \\ \hline

		\begin{minipage}{0.49\linewidth}
		\vspace{1mm}
		\centering
		\includegraphics[width=\linewidth,trim={1.5cm 1cm 1.5cm 1cm}, clip]{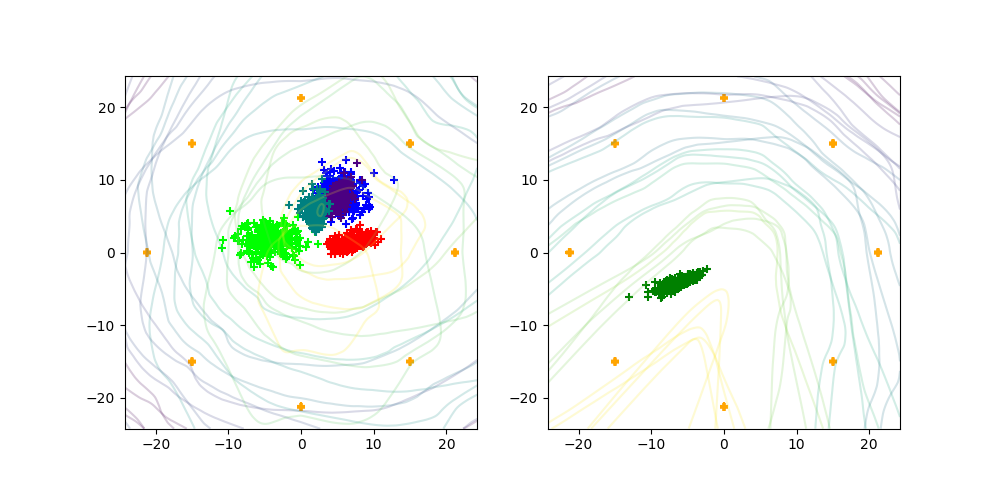} 
		\textit{\small Iteration 10}	
		\end{minipage} &

		\begin{minipage}{0.49\linewidth}
		\vspace{1mm}
		\centering
		\includegraphics[width=\linewidth,trim={1.5cm 1cm 1.5cm 1cm}, clip]{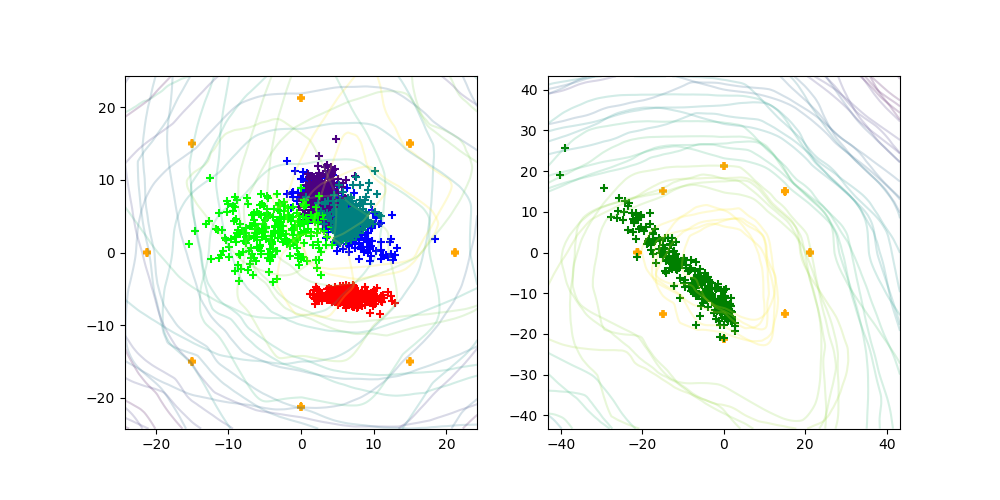} 
		\textit{\small Iteration 15}	\vspace{1em}
		\end{minipage} \\ \hline

		\begin{minipage}{0.49\linewidth}
		\vspace{1mm}
		\centering
		\includegraphics[width=\linewidth,trim={1.5cm 1cm 1.5cm 1cm}, clip]{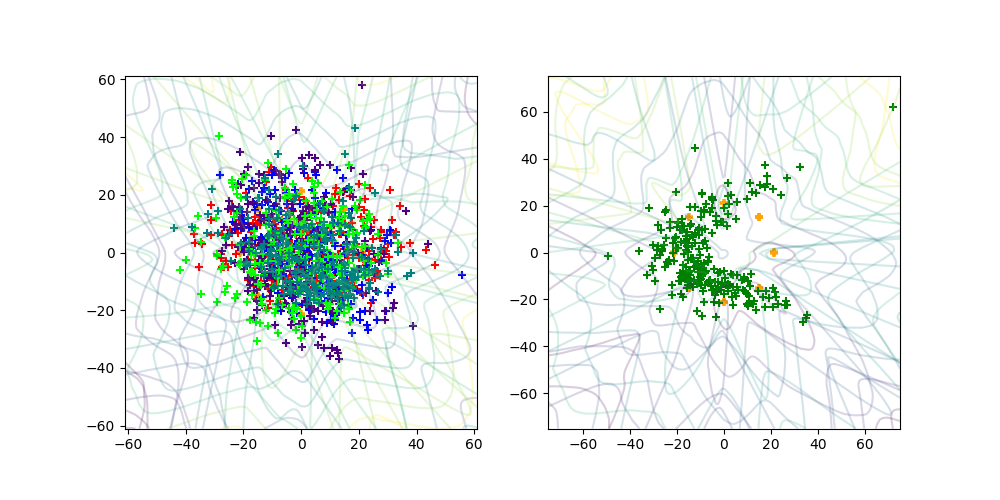} 
		\textit{\small Iteration 60}	\vspace{1em}
		\end{minipage} &

		\begin{minipage}{0.49\linewidth}
		\vspace{1mm}
		\centering
		\includegraphics[width=\linewidth,trim={1.5cm 1cm 1.5cm 1cm}, clip]{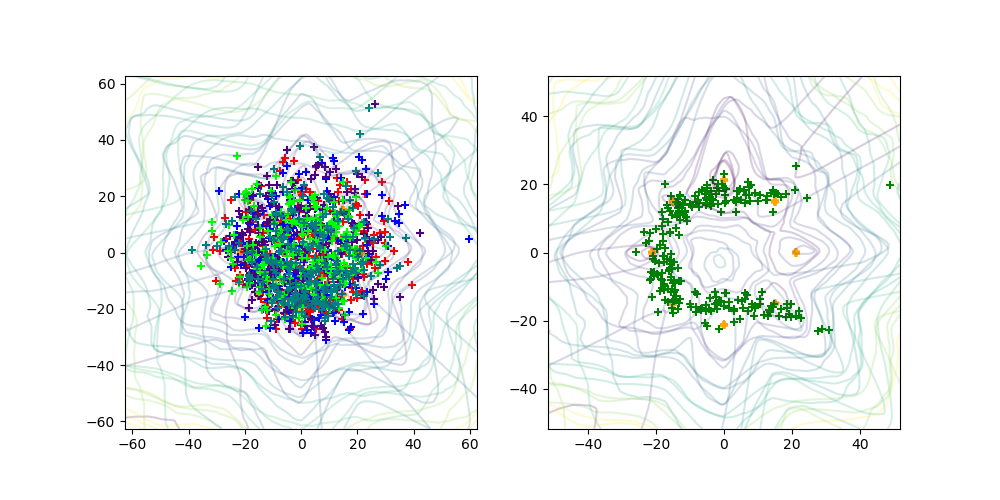} 
		\textit{\small Iteration 80}	\vspace{1em}
		\end{minipage} \\ \hline

		\begin{minipage}{0.49\linewidth}
		\vspace{1mm}
		\centering
		\includegraphics[width=\linewidth,trim={1.5cm 1cm 1.5cm 1cm}, clip]{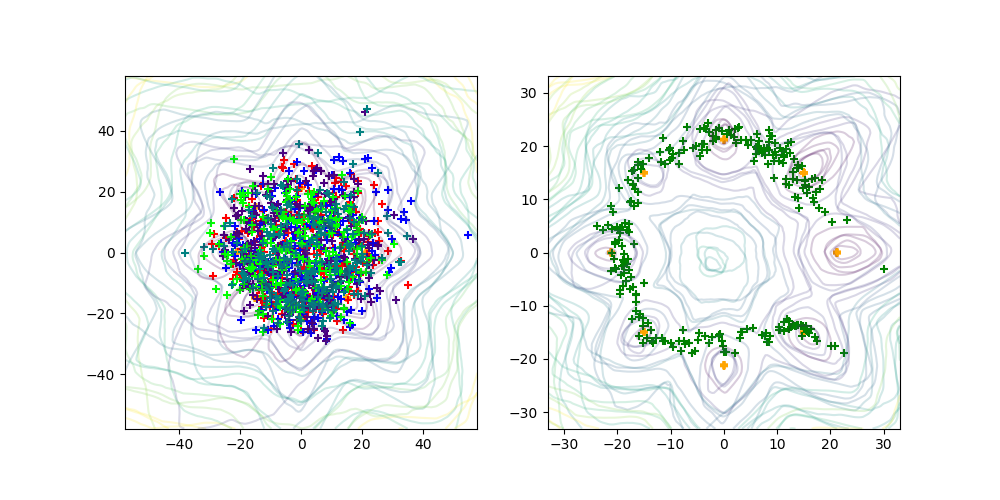}  
		\textit{\small Iteration 100}	\vspace{1em}
		\end{minipage} &

		\begin{minipage}{0.49\linewidth}
		\vspace{1mm}
		\centering
		\includegraphics[width=\linewidth,trim={1.5cm 1cm 1.5cm 1cm}, clip]{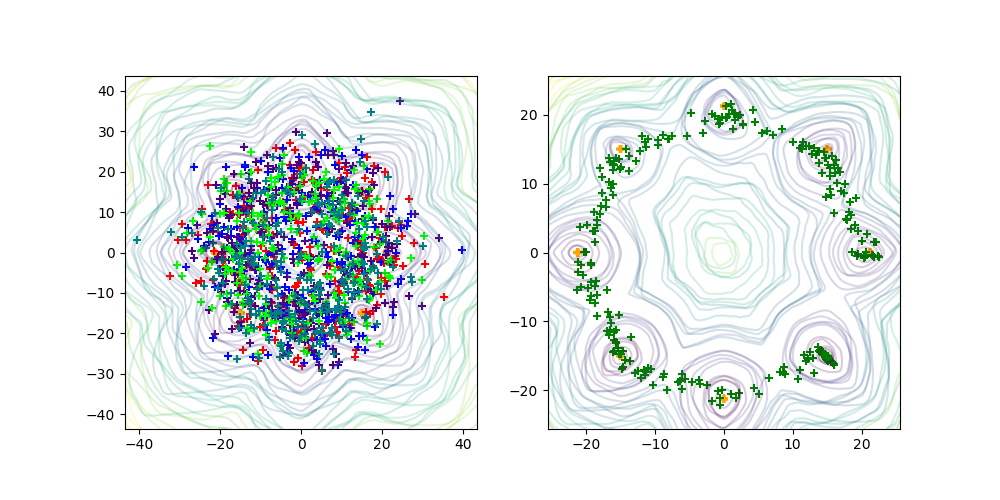} 
		\textit{\small Iteration 300}	 \vspace{1em}
		\end{minipage}
\end{tabular}
	\caption{\textbf{5-S-WGAN} experiment on the \textbf{8-GMM} toy dataset (best seen in color).
	Real data samples are illustrated in orange.
	In each image pair, we illustrate samples from the five local generators and from the global generator, on the left (in separate color) and on the right (in green), respectively.	
	The displayed contours represent the level sets of the discriminators $\mathcal{D}$ and $\mathcal{D}^{msg}$--illustrated on the left and  right of each image pair, respectively, where yellow is low and purple is high.
	}
	\label{fig-exp_p5-1}
\end{figure*}

In SGAN the global generator can also be updated after each update of any of the local pairs. In our preliminary results this did \textbf{not} hurt the performance. We illustrate such experiment in Figure~\ref{fig-exp_p5-1}. On the other hand, in classical training imbalanced frequency of the parameters' updates between the generator and the discriminator may cause failure in convergence.
Nonetheless, for a fair comparison in \textit{all the experiments herein (including the real data experiments)} at each iteration of SGAN we update the parameters of the global generator as many times as the parameters of any local generator have been updated. We recall that, the difference is that the global generator is trained jointly versus the ''messenger`` discriminators. 

\section{Experiments on real-world datasets (an extension)}\label{app:sgan_exp_real}
\subsection{Details on the implementation}
We did experiments in two set-ups: 
\begin{enumerate*}[series = tobecont, itemjoin = \quad, label=(\roman*)]
\item using separate $2{\cdot}(N{+}1)$ networks; as well as
\item using sharing parameters among the networks.
\end{enumerate*}
In the latter case, approximately half of the parameters of each network are shared among the corresponding other $N$ networks (discriminators or generators).
For clarity herein for SGAN with separate networks we use the standard prefix of \textbf{N-S-}, and for SGAN with weight sharing we use prefix \textbf{N-SW-}, where N denotes the number of local pairs being used (see \S~\ref{par-datasets}).

For some variants of the GAN algorithm, we observe that sharing parameters may lead to a small difference between samples from the  local generators.
In these cases, \textit{SGAN training is as good or marginally better} compared to regular training. Hence, for \textit{DCGAN in particular, we recommend using separate networks (rather than weight sharing)}.

We used learning rate of $1{\cdot}10^{-5}$, and a batch size of $50$ and $64$ for (FASHION)MNIST and the rest of the datasets, respectively. 
Unless otherwise stated, we used the \textit{Adam} optimizer~\cite{adam} whose hyperparameters (one parameter used for computing running averages of gradient and another for its square) we fixed to $0.5$ and $0.999$, as in~\cite{dcgan}.

\paragraph{Implementation of the experiments on image datasets.}
For \textbf{MNIST} we did experiments using both MLPs and CNNs for the generators and the discriminators.
In the former case, the architectures were almost identical to those used for the toy experiments, except that the first layer was adjusted for input space of $28{\times}28$.
In the latter case, we used input space of $28{\times}28$ and we started with the DCGAN implementation~\cite{dcgan} and changed it accordingly to the input space.
In particular, we reduced the number of  2D transposed convolution layers from $5$ to $4$ and adjusted the hidden layers' sizes accordingly to the dimensions used for the real data space.

For \textbf{CIFAR10} unless otherwise emphasized, we used $32{\times}32$ image space.
For the rest of the image datasets--unless otherwise stated, we used $64{\times}64$ input space and the original DCGAN~\cite{dcgan} architecture, as provided by the authors. The implementation of DCGAN ~\cite{dcgan} uses Batch Normalization layers~\cite{bnorm}.

\paragraph{Implementation of the experiments on one Billion Word Benchmark.}
We started from the provided implementation of~\cite{wgangp} and implemented our method.
In particular, the character-level generative language model is implemented as a $1D$ CNN using $4$ ResNet blocks~\cite{resnet},
which network maps a latent vector into a sequence of one-hot character vectors of dimension 32.
The discriminator is also a $1D$ CNN, that takes as input sequences of such character embeddings of size 32.

As optimization method we used \textit{RMSProp}~\cite{rmsprop}.

\paragraph{Separate networks.}\label{sec:real-separate}

\begin{figure}[!htb]
\centering
\includegraphics[width=\linewidth]{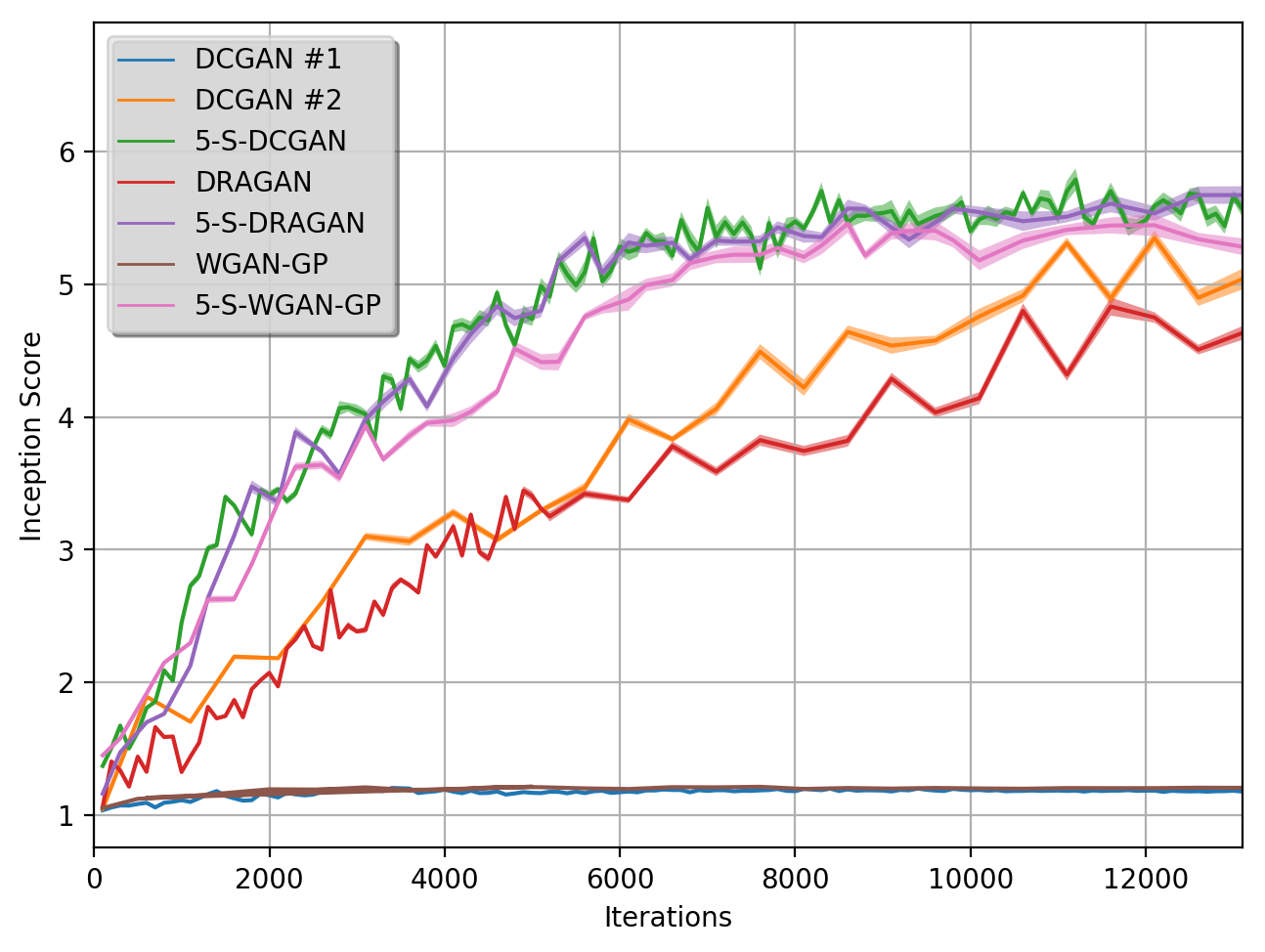}
\caption{
Inception Score~\cite{Salimans2016improvingGANs} on \textbf{CIFAR10} (see text \S~\ref{sec:real-separate}). Best seen in color.}
\label{fig:is_cifar}
\end{figure}

\begin{figure}[!htb]
\centering
\includegraphics[width=\linewidth]{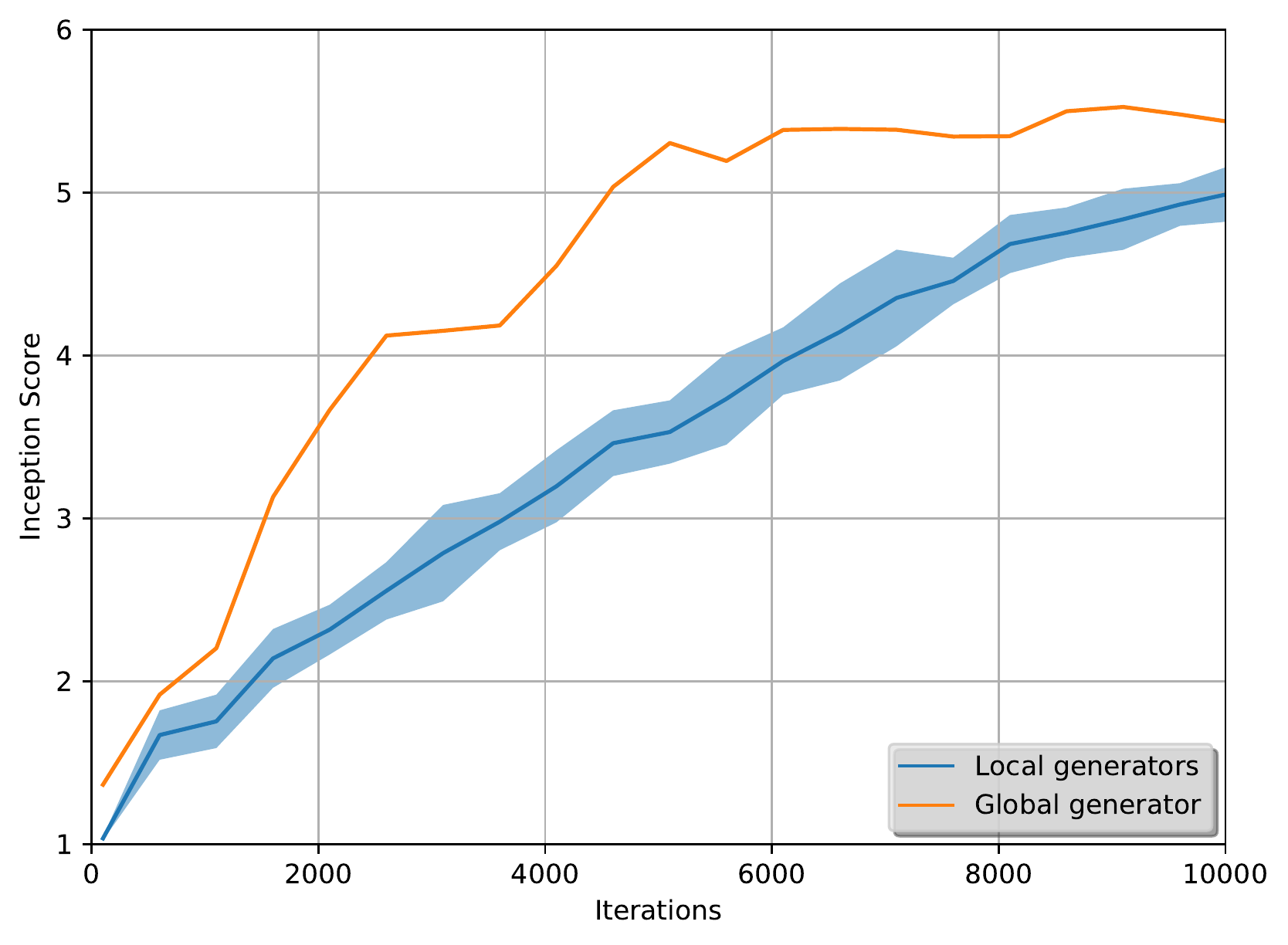}
\caption{\textbf{10-S-DCGAN}, on on \textbf{CIFAR10}, using real data space of $32{\times}32$ (best seen in color). We plot the Inception Score~\cite{Salimans2016improvingGANs} of the global generator (orange) as well as the scores of the local generators (blue).}
\label{fig:is_cifar_local_vs_global}
\end{figure}

\begin{figure}[!htb]
\centering
\begin{tabular}{c@{\hskip 0.02in}c}
\includegraphics[width=.48\linewidth]{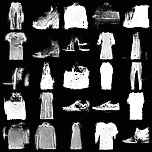} &
\includegraphics[width=.48\linewidth]{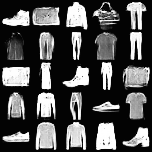} \\
\end{tabular}
\caption{
Samples of \textbf{DCGAN} and \textbf{5-S-DCGAN} on \textbf{FASHION-MNIST} taken at the $6000$\textit{-th} iteration, on the left and right, respectively. Using image sizes of $28{\times}28$.}
\label{fig:sgan_fashion}
\end{figure}

\begin{figure}[!htb]
\begin{tabular}{c@{\hskip 0.02in}c@{\hskip 0.02in}c}
	\begin{minipage}{0.3\linewidth}
		\begin{tabular}{c}
		\includegraphics[width=\linewidth]{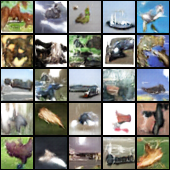}\\
		\textit{\small Global generator}
		\end{tabular}	
	\end{minipage} 	&
	\begin{minipage}{0.3\linewidth}
		\begin{tabular}{c}
		\includegraphics[width=\linewidth]{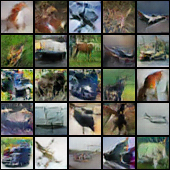} \\
		\textit{\small Local generator \#1}	\\
		\end{tabular}	
	\end{minipage} 	&
	\begin{minipage}{0.3\linewidth}
		\begin{tabular}{c}
		\includegraphics[width=\linewidth]{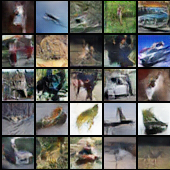} \\
		\textit{\small Local generator \#2}	\\
		\end{tabular}	
	\end{minipage} \hspace{.2cm} 	\vspace{.09in} \\ \noindent 
	\begin{minipage}{0.3\linewidth}
		\begin{tabular}{c}
		\includegraphics[width=\linewidth]{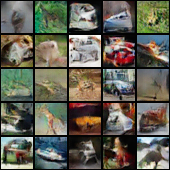} \\
		\textit{\small Local generator \#3}	\\
		\end{tabular}	
	\end{minipage} 	&
	\begin{minipage}{0.3\linewidth}
		\begin{tabular}{c}
		\includegraphics[width=\linewidth]{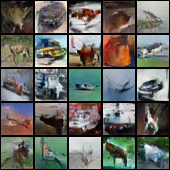} \\
		\textit{\small Local generator \#4}
		\end{tabular}	
	\end{minipage} 	&
	\begin{minipage}{0.3\linewidth}
		\begin{tabular}{c}
		\includegraphics[width=\linewidth]{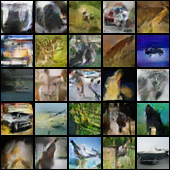}\\
		\textit{\small Local generator \#5}	
		\end{tabular}	
	\end{minipage} \hspace{.3cm} 	
\end{tabular}
	\caption{\textbf{5-S-DRAGAN} on \textbf{CIFAR10} at $40{\cdot}10^3$\textit{-th} iteration, and $32{\times}32$ real data space.}
	\label{fig-5s_dragan-cifar10}
\end{figure}

\begin{figure}[!htb]
\begin{tabular}{c@{\hskip 0.02in}c@{\hskip 0.02in}c}
	\begin{minipage}{0.3\linewidth}
		\begin{tabular}{c}
		\includegraphics[width=\linewidth]{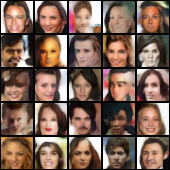}\\
		\textit{\small Global generator}
		\end{tabular}	
	\end{minipage} 	&
	\begin{minipage}{0.3\linewidth}
		\begin{tabular}{c}
		\includegraphics[width=\linewidth]{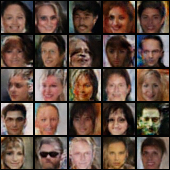} \\
		\textit{\small Local generator \#1}	\\
		\end{tabular}	
	\end{minipage} 	&
	\begin{minipage}{0.3\linewidth}
		\begin{tabular}{c}
		\includegraphics[width=\linewidth]{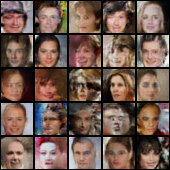} \\
		\textit{\small Local generator \#2}	\\
		\end{tabular}	
	\end{minipage} \hspace{.2cm} 	\vspace{.09in} \\ \noindent 
	\begin{minipage}{0.3\linewidth}
		\begin{tabular}{c}
		\includegraphics[width=\linewidth]{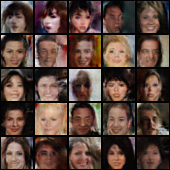} \\
		\textit{\small Local generator \#3}	\\
		\end{tabular}	
	\end{minipage} 	&
	\begin{minipage}{0.3\linewidth}
		\begin{tabular}{c}
		\includegraphics[width=\linewidth]{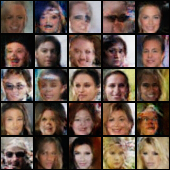} \\
		\textit{\small Local generator \#4}
		\end{tabular}	
	\end{minipage} 	&
	\begin{minipage}{0.3\linewidth}
		\begin{tabular}{c}
		\includegraphics[width=\linewidth]{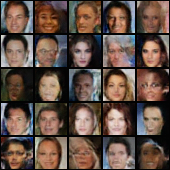}\\
		\textit{\small Local generator \#5}	
		\end{tabular}	
	\end{minipage} \hspace{.3cm} 	
\end{tabular}
	\caption{\textbf{5-S-DCGAN} on \textbf{CelebA} at $1{\cdot}10^3$\textit{-th} iteration, and $32{\times}32$ real data space..}
	\label{fig-5s_dcgan-celeba}
\end{figure}

\begin{figure}[h!]
\centering
\begin{tabular}{c@{\hskip 0.02in}c}
	\includegraphics[width=.45\linewidth]{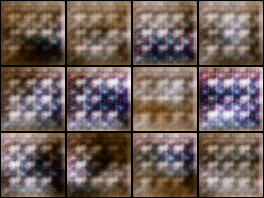} &
	\includegraphics[width=.45\linewidth]{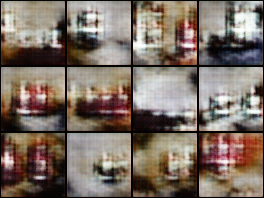} 
    \vspace{0.02in} \\
	\includegraphics[width=.45\linewidth]{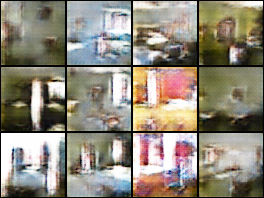} &
	\includegraphics[width=.45\linewidth]{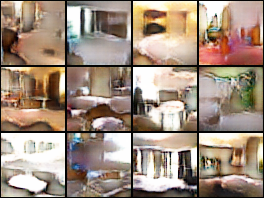} 
	\vspace{0.02in} 	\\ 
	\includegraphics[width=.45\linewidth]{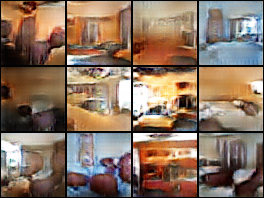} &
	\includegraphics[width=.45\linewidth]{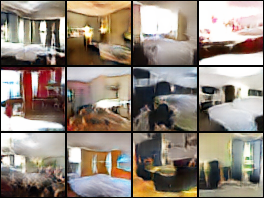} 
	\vspace{0.02in} \\ 
	\includegraphics[width=.45\linewidth]{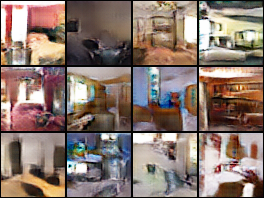} &
	\includegraphics[width=.45\linewidth]{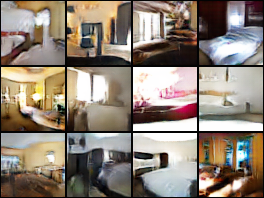} \hspace{.01\linewidth}\\
	\textbf{\small DRAGAN}  		& 		\textbf{\small 5-SW-DRAGAN} \\ 
\end{tabular}
	\caption{\textbf{DRAGAN} and \textbf{5-SW-DRAGAN} on \textbf{LSUN-bedroom} at the $1000$\textit{-th}, $5{\cdot}10^3$\textit{-th}, $10{\cdot}10^3$\textit{-th} and $14{\cdot}10^3$\textit{-th} iteration, from top to bottom row, respectively. Using $64{\times}64$ real data space.}
	\label{fig-5sw_dragan-lsun}
\end{figure}

In Figure~\ref{fig:is_cifar} we plot the Inception Score~\cite{Salimans2016improvingGANs} using its original implementation in TensorFlow~\cite{tensorflow}.
To avoid any difference, for \latin{e.g.} the different sampling of the dataset while training, \textit{for any regular training--of one pair, we run a \textbf{separate} experiment}, rather than calculating the inception scores of the local pairs.
For \textbf{DCGAN \#1} although from the generated samples we could observe that the algorithm is converging, the Inception Score was low.
For a fair comparison, we re-run the experiment while varying the hyperparameters, denoted as \textbf{DCGAN \#2} in Figure~\ref{fig:is_cifar}.
In Figure~\ref{fig:is_cifar_local_vs_global} we show the Inception scores of the  global generator and the local generators.

In Figure~\ref{fig:sgan_fashion} we show samples of \textbf{5-S-DCGAN} on \textbf{FASHION-MNIST} (on the right), as well as of \textbf{DCGAN} (on the left). Note that, the latter experiment was done separately.
Figures~\ref{fig-5s_dragan-cifar10} \&~\ref{fig-5s_dcgan-celeba} depict samples using \textbf{DRAGAN} and \textbf{DCGAN}, respectively. We see that the global generator converges much earlier then the local ones.


\paragraph{Shared parameters.}
In the sequel, the illustrated samples are taken from generators that share their parameters. Note that the samples from regular training are always obtained by a separate experiment (in contrast to taking samples from the local pairs), due to the weight sharing in SGAN.

In Figure~\ref{fig-5sw_dragan-lsun} we show samples when training \textbf{DRAGAN} and \textbf{5-SW-DRAGAN} on \textbf{LSUN-bedroom} with input space of $64{\times}64$.
Finally, in Figure~\ref{fig-nlp} we show samples when training on the \textbf{ Billion Word } dataset.

\noindent
\begin{figure}[h!]
\centering
\noindent
	\begin{minipage}{.45\linewidth}
	\centering
\begin{Verbatim}[fontsize=\tiny, frame=single]
eeeeeeeeeeeeeeeeeeeeeeeeeeeeeeee
eeeeeeeeeeeeeeeeeeeeeeeeeeeeeeee
eeeeeeeeeeeeeeeeeeeeeeeeeeeeeeee
eeeeeeeeeeeeeeeeeeeeeeeeeeeeeeee
eeeeeeeeeeeeeeeeeeeeeeeeeeeeeeee
eeeeeeeeeeeeeeeeeeeeeeeeeeeeeeee
eeeeeeeeeeeeeeeeeeeeeeeeeeeeeeee
eeeeeeeeeeeeeeeeeeeeeeeeeeeeeeee
eeeeeeeeeeeeeeeeeeeeeeeeeeeeeeee
eeeeeeeeeeeeeeeeeeeeeeeeeeeeeeee
eeeeeeeeeeeeeeeeeeeeeeeeeeeeeeee
\end{Verbatim}
	\end{minipage}  
	\begin{minipage}{.45\linewidth}
	\begin{Verbatim}[fontsize=\tiny, frame=single]
The Larntare F bucnt 1h setirder
The tielirc indian on orabo stir
We dola wan Fobkbomn hrcas and 8
Tod letrocsix r telt car rntr ce
Then on s indent on leWand ghis 
Ja candt conteltirt in ald do ce
Gaid nochir Weabsilan of ansany
The ticosgose arc on Mesinntelat
More wucarcanosnped rochusroe t`
They derato raEyand soalceatecst
De Ths chsrc s aeareP thel ea t 
\end{Verbatim}
	\end{minipage}\\	
\noindent
	\begin{minipage}{.45\linewidth}
	\centering
\begin{Verbatim}[fontsize=\tiny, frame=single]
" S4vvvoFlnls anr ans ffrcinns s
Thon taa forinint ssroso siwfd f
Hothst bffld 'nvlonyoiar" cov sh
Woisi'n fof Monisg dhak N`f fnv 
ThD fas ong fafpn so n wns is of
D" wiay wyd alvriMbnlor M nld ff
DDd4y vooc onl vocfay w s offo f
" c4Df co đ noy soonlono ans war
ln wns bfrfncorfiw Thofv lnnd fo
Dy wad Dld at N fovl dcy fot aor
ThD doDn bacd d vffnonlo anfofin
Th' toits isg thid st hsgo ffffs
" coracgod Mfopf đlny thisg  aff
\end{Verbatim}

	\end{minipage}  
	\begin{minipage}{.45\linewidth}
	\begin{Verbatim}[fontsize=\tiny, frame=single]
The conareed same ming tay spid 
Then the gioncolly the can id co 
She  beme lant arecong nelode .  
It taecanting the they trehos so 
In the later asterol antarlist n  
Sion ot tndy ttin an os pomcerer
The pither canned Sblets castery 
They BastitiBented tome man angu 
I lant taot suncedrthet prourpli 
And Biecon eels acecount tre Car 
This tain Datertlals comegel yan 
Is the Woilg  ate costort thab f 
Thin Incin is Inlasar cumand mot
\end{Verbatim}
	\end{minipage}
\caption{
Snippets from \textbf{WGAN} (left) and \textbf{5-SW-WGAN} (right) on the  \textbf{One Billion Word Benchmark},
taken at the $700$\textit{-th} and $2500$\textit{-th} iteration (top and bottom row, respectively).
}
\label{fig-nlp}
\end{figure}

\checknbdrafts

\end{document}